\documentclass[11pt, letterpaper, logo, onecolumn, copyright]{main}

\usepackage[authoryear, sort&compress, round]{natbib}

\usepackage[inkscapeformat=png]{svg}

\usepackage[most, breakable, skins]{tcolorbox}

\tcbuselibrary{skins}
\usepackage{lipsum}
\usepackage{tabularx}
\usepackage{afterpage}
\usepackage{booktabs}
\usepackage{subcaption}
\usepackage{makecell}
\usepackage{multirow}
\usepackage{bm}
\usepackage{multicol}
\usepackage{array}
\usepackage{float}
\usepackage{listings, listings-rust}
\usepackage{fontawesome5}
\usepackage{hyperref}
\usepackage{amssymb,graphicx}
\usepackage[dvipsnames]{xcolor}
\usepackage{cleveref}
\usepackage{longtable}
\usepackage{pdflscape}
\usepackage{adjustbox}
\usepackage{nicematrix}
\usepackage{CJKutf8}
\usepackage{ragged2e}
\usepackage{colortbl}
\usepackage{enumitem}
\usepackage[ruled,linesnumbered]{algorithm2e}
\usepackage{pifont}
\usepackage[htt]{hyphenat}
\usepackage{amsmath}
\usepackage{amsthm}  
\usepackage{mathrsfs} 
\usepackage{circledsteps}
\usepackage{diagbox}
\usepackage{bookmark}

\newtheorem{definition}{Definition}
\newtheorem{lemma}{Lemma}

\usepackage{wrapfig}
\usepackage{xspace}
\usepackage{tikz}
\usepackage[normalem]{ulem}

\usepackage{pgfplots}
\usepackage{pgfplotstable}
\pgfplotsset{compat=1.18}

\definecolor{medgray55}{gray}{0.55}
\definecolor{medgray}{gray}{0.7}
\definecolor{litegray}{gray}{0.9}
\definecolor{gblue}{RGB}{210, 227, 252}
\definecolor{gred}{RGB}{250, 210, 207}
\definecolor{gyellow}{RGB}{254, 239, 195}
\definecolor{ggreen}{RGB}{206, 234, 214}
\definecolor{gorange}{RGB}{254, 223, 200}

\definecolor{gblue9}{RGB}{23, 78, 166}
\definecolor{gred9}{RGB}{165, 14, 14}
\definecolor{gyellow9}{RGB}{227, 116, 0}
\definecolor{ggreen9}{RGB}{13, 101, 45}
\definecolor{gorange9}{RGB}{176, 96, 0}

\definecolor{myblue}{rgb}{0,0,1}
\definecolor{myred}{rgb}{1,0,0}
\definecolor{mylightgray}{gray}{0.95}
\definecolor{myCite}{HTML}{1C4587}

\definecolor{highlightblue}{HTML}{185ABC}
\definecolor{cellHighlight}{HTML}{dbefff}

\usepackage{minitoc}

\noptcrule


\newcolumntype{L}[1]{>{\raggedright\let\newline\\\arraybackslash\hspace{0pt}}m{#1}}
\newcolumntype{C}[1]{>{\centering}m{#1}}

\newcolumntype{R}[1]{>{\raggedleft\let\newline\\\arraybackslash\hspace{0pt}}m{#1}}

\definecolor{ao}{rgb}{0.0, 0.0, 1.0}

\newcommand\vcent[1]{\vcenter{\hbox{#1}}}

\newcommand\loudspeaker[1][3]{\ensuremath{\vcent{\rule{.6ex}{.6ex}}\kern-.5ex
  \vcent{\scalebox{.6}[1]{\rotatebox[origin=center]{90}{$\blacktriangle$}}}
  \ifnum#1>0\relax\kern.05ex\vcent{\scalebox{.4}{\ttfamily)}}
  \ifnum#1>1\relax\kern-.4ex\vcent{\scalebox{.56}{\ttfamily)}}
  \ifnum#1>2\relax\kern-.55ex\vcent{\scalebox{.7}{\ttfamily)}}
  \fi\fi\fi}
}

\makeatletter
\renewcommand\subparagraph{
 \@startsection {subparagraph}{5}{\z@ }{3.25ex \@plus 1ex
 \@minus .2ex}{-1em}{\normalfont \normalsize \bfseries }}
\makeatother

\let\cite\citep
\hypersetup{
  citecolor = myCite,  
  linkcolor = myCite,   
  urlcolor  = myCite
}

\title{BAPO: Stabilizing Off-Policy Reinforcement Learning for LLMs via Balanced Policy Optimization with Adaptive Clipping}
\author{
    Zhiheng Xi$^1$$^{* \dag}$,  Xin Guo$^1$$^*$, Yang Nan$^1$, Enyu Zhou$^1$, Junrui Shen$^1$,  \\
\textbf{Wenxiang Chen$^1$, Jiaqi Liu$^1$, Jixuan Huang$^1$, Zhihao Zhang$^1$, Honglin Guo$^1$,} \\
\textbf{Xun Deng$^2$, Zhikai Lei$^2$, Miao Zheng$^2$, Guoteng Wang$^2$, Shuo Zhang$^2$, Peng Sun$^2$,} \\
\textbf{ Rui Zheng$^2$, Hang Yan$^2$, Tao Gui$^{1,3}$$^\dag$, Qi Zhang$^1$$^\dag$, Xuanjing Huang$^1$}
\\
$^1$Fudan University $^2$Shanghai Qiji Zhifeng Co., Ltd. $^3$Shanghai Innovation Institute \\
\texttt{zhxi22@m.fudan.edu.cn, \{tgui,qz\}@fudan.edu.cn} 
}
\vspace{-50pt}
\begin{abstract}
Reinforcement learning (RL) has recently become the core paradigm for aligning and strengthening large language models (LLMs). Yet, applying RL in off-policy settings–where stale data from past policies are used for training–improves sample efficiency, but remains challenging: policy entropy declines sharply, optimization often becomes unstable and may even collapse. Through theoretical and empirical analysis, we identify two key insights: (i) \textbf{an imbalance in optimization}, where negative-advantage samples dominate the policy gradient, suppressing useful behaviors and risking gradient explosions; and (ii) \textbf{the derived Entropy-Clip Rule}, which reveals that the fixed clipping mechanism in PPO-like objectives systematically blocks entropy-increasing updates, thereby driving the policy toward over-exploitation at the expense of exploration. Building on these insights, we propose \textbf{BA}lanced \textbf{P}olicy \textbf{O}ptimization with Adaptive Clipping (\textbf{BAPO}), a simple yet effective method that dynamically adjusts clipping bounds to adaptively re-balance positive and negative contributions, preserve entropy, and stabilize RL optimization. Across diverse off-policy scenarios–including sample replay and partial rollout–BAPO achieves fast, stable, and data-efficient training. On AIME 2024 and AIME 2025 benchmarks, our 7B BAPO model surpasses open-source counterparts such as SkyWork-OR1-7B, while our 32B BAPO model not only achieves state-of-the-art results among models of the same scale but also outperforms leading proprietary systems like o3-mini and Gemini-2.5-Flash-Thinking.
\end{abstract}

\begin{document}

\doparttoc
\faketableofcontents

\begingroup
  \renewcommand\thefootnote{}
  \footnote{\textsuperscript{*}Equal contribution.
            \textsuperscript{\dag}Corresponding authors.}
    \footnote{\textsuperscript{1}Our code are available at \url{https://github.com/WooooDyy/BAPO}.}
  \addtocounter{footnote}{-1}
\endgroup

\vspace{-30pt}
\maketitle

\begin{figure}[!ht]
\begin{center}
\vspace{-10pt}
\includegraphics[width=0.9\linewidth]{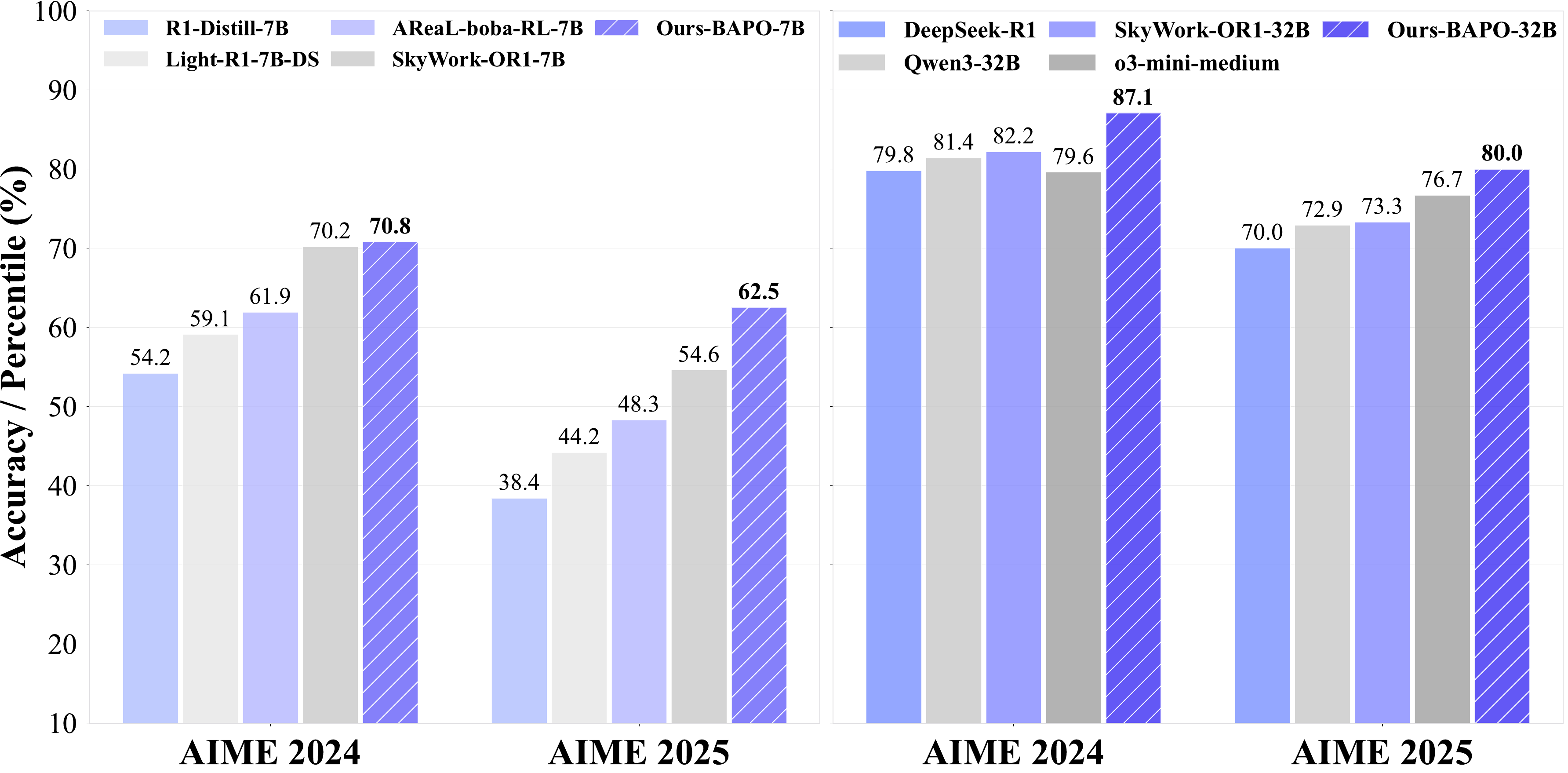}
\end{center}
\vspace{-16pt}
\caption{Performance of BAlanced Policy Optimization with Adaptive Clipping (BAPO).}
\label{fig:results_r1}

\end{figure}

\section{Introduction}

Reinforcement learning (RL) has become a pivotal paradigm for optimizing large language models (LLMs) \citep{zhang2025survey}, delivering significant improvements in complex tasks such as reasoning \citep{openai_o1, DeepSeek-R1}, coding \citep{claude_code}, and agentic decision-making \citep{DBLP:journals/corr/abs-2507-20534}. Among RL methods, off-policy RL–where the rollout policy (behavior policy) differs from the training policy (target policy)–emerges as particularly promising \citep{DBLP:journals/corr/abs-2503-14286, DBLP:journals/corr/abs-2506-20520}. It offers high sample efficiency and tolerance to data staleness, making it well-suited for extremely long-horizon and challenging scenarios, while also aligning more naturally with features in modern AI infrastructures such as partial rollout \citep{DBLP:journals/corr/abs-2501-12599, DBLP:journals/corr/abs-2505-24298}.

\begin{figure}[htbp]
\begin{center}
\includegraphics[width=0.98\linewidth]{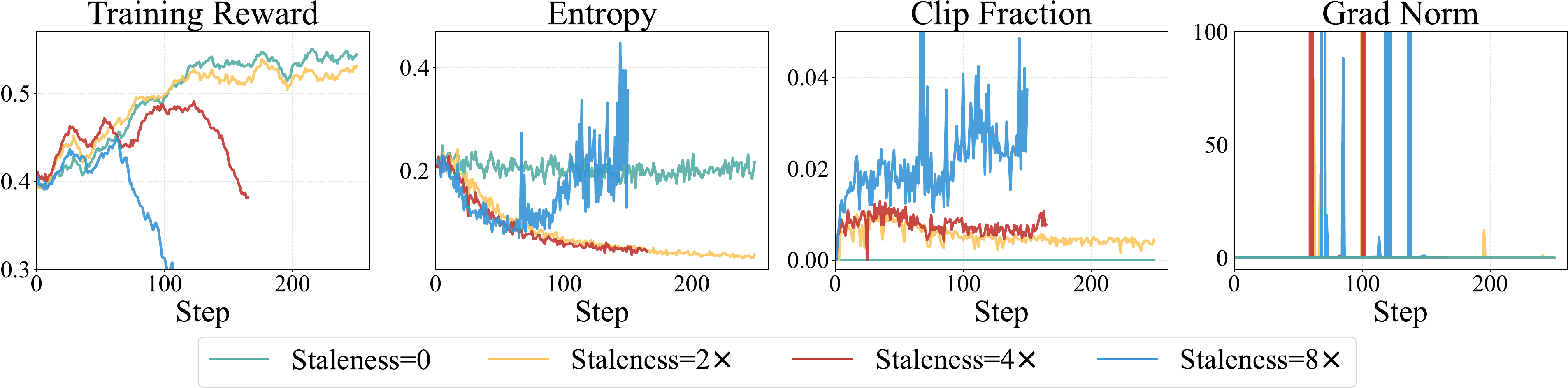}
\end{center}
\vspace{-10pt}
\caption{
Preliminary results with different data staleness. As the staleness increases, the model suffers from unstable optimization, decreasing entropy, and even a sudden collapse in training.
}
\label{fig:on_off_policy}
\end{figure}

However, applying off-policy RL to LLMs introduces substantial challenges \citep{DBLP:journals/corr/abs-2503-14476, DBLP:journals/corr/abs-2506-20520}. As shown in Figure~\ref{fig:on_off_policy}, increasing data staleness leads to unstable optimization, exploding gradient and even collapse. Meanwhile, policy entropy declines sharply, reflecting reduced exploratory capacity and a bias toward over-exploitation. By contrast, on-policy training–where rollout and target policies coincide–remains stable across metrics, consistent with prior studies \citep{DBLP:journals/corr/abs-2405-08448, DBLP:journals/corr/abs-2503-14286, DBLP:journals/corr/abs-2506-20520}.

To understand the instability of off-policy training, we conduct a comprehensive theoretical and empirical analysis to reveal two key insights.
We first demonstrate an \textbf{imbalance in optimization}: policy updates are often dominated by negative-advantage samples, producing excessive penalization signals that suppress even neutral or correct actions and may cause gradient explosions \citep{DBLP:journals/corr/abs-2308-08998}. We then derive and empirically validate the \textbf{Entropy-Clip Rule} in the widely-used PPO \citep{DBLP:journals/corr/SchulmanWDRK17} and GRPO \citep{DBLP:journals/corr/abs-2402-03300e}, showing that the clipping mechanism in PPO-like objectives blocks many low-probability positive tokens while over-penalizing low-probability negatives. This systematically excludes entropy-increasing updates, sharpens the output distribution, and drives policies toward over-exploitation at the cost of exploration. 

\begin{figure}[t]
\begin{center}
\includegraphics[width=0.9\linewidth]{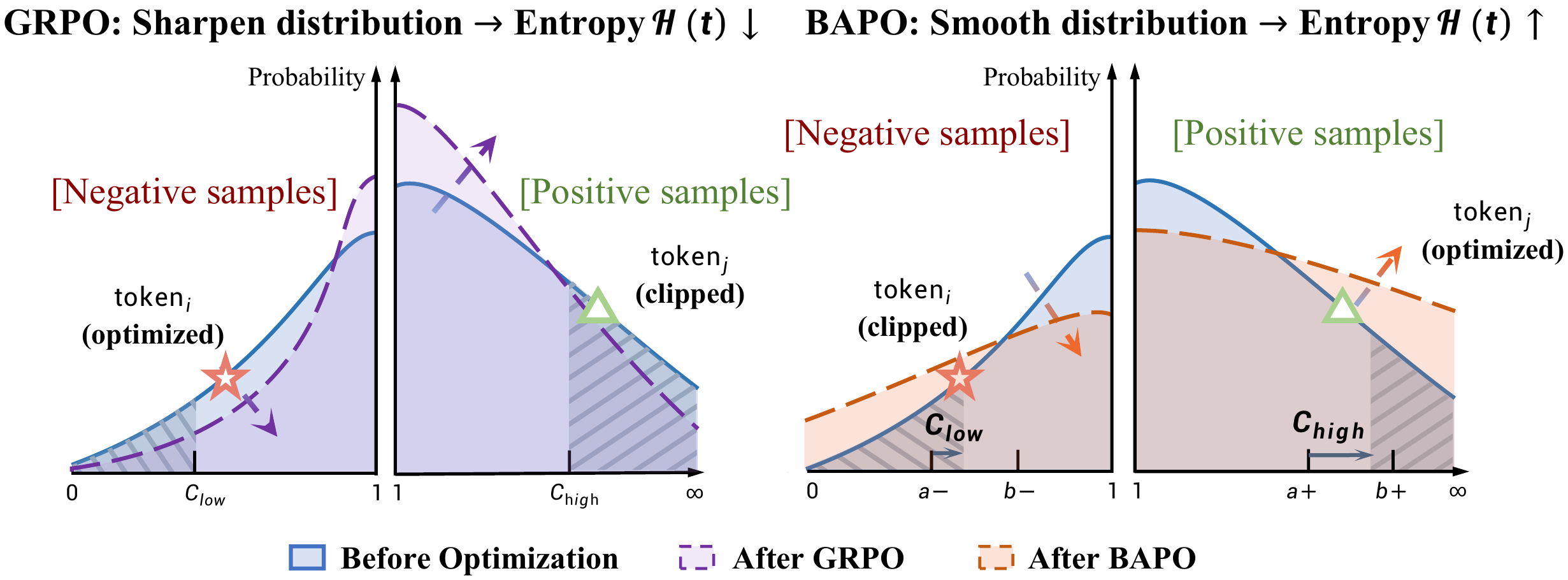}
\end{center}
\vspace{-5pt}
\caption{An illustration of our proposed BAPO. \textbf{(Left)} Baseline methods like GRPO use symmetric fixed clipping bounds, reinforcing high-probability positive tokens while penalizing excessive low-probability negatives, leading to sharp distributions and entropy collapse. \textbf{(Right)} BAPO dynamically adjusts the clipping bounds $c_\textnormal{low}$ and $c_\textnormal{high}$ based on the loss contributions from positive tokens. It excludes overly negative tokens \smash{\raisebox{-0.2\baselineskip}{\includegraphics[height=\baselineskip]{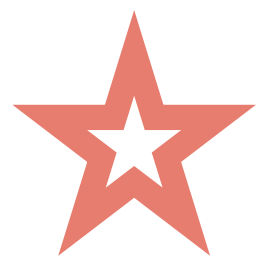}}} to maintain a smoother distribution and incorporates previously clipped positive tokens \smash{\raisebox{-0.2\baselineskip}{\includegraphics[height=\baselineskip]{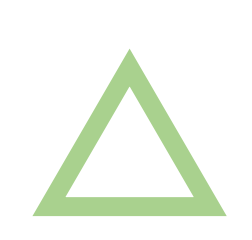}}} to preserve entropy balance.}
\label{fig:main}
\end{figure}

Based on these insights, we propose \textbf{BA}lanced \textbf{P}olicy \textbf{O}ptimization with Adaptive Clipping (\textbf{BAPO}), a new method for stable and effective off-policy RL. BAPO dynamically adjusts the clipping bounds to re-balance positive and negative contributions for each update step, incorporate low-probability positives while filtering excessive negatives, and preserve policy entropy–achieving a better balance between exploration and exploitation.
An overview of our approach is illustrated on the right side of Figure~\ref{fig:main}. 

Experiments across diverse off-policy scenarios–including sample replay, partial rollout, and varying degrees of staleness–on base models such as DeepSeek-R1-Distill-Qwen-7B \citep{DeepSeek-R1} and OctoThinker-Llama3.2-3B-Long-Zero \citep{DBLP:journals/corr/abs-2506-20512} show that BAPO consistently yields significant improvements. Our 7B model achieves scores of $70.8$ on AIME24 and $62.5$ on AIME25, surpassing open-source counterparts such as SkyWork-OR1-7B \citep{skywork_or1}. Moreover, our 32B model reaches $87.1$ on AIME24 and $80.0$ on AIME25, outperforming both comparably scaled open-source models like Qwen3-32B \citep{DBLP:journals/corr/abs-2505-09388} and leading proprietary systems including o3-mini-medium \citep{openai_o3} and Gemini-2.5-Flash-Thinking \citep{DBLP:journals/corr/abs-2507-06261}.

Our contributions are summarized as follows:
\begin{itemize}[leftmargin=*]
\item We identify and analyze two key insights behind instability in off-policy RL for LLMs: the {imbalanced optimization} and the {Entropy-Clip Rule}. (\S \ref{sec:analysis})
\item We propose BAPO, a new RL algorithm that dynamically adjusts clipping bounds to balance positive and negative signals, preserving entropy for exploration, and stabilizing training. (\S \ref{sec:methods})
\item We validate BAPO across multiple backbones, model scales, and off-policy settings, showing that it achieves stable optimization and competitive results with proprietary systems.  (\S \ref{sec:experiments})
\end{itemize}

\section{Preliminaries}
\subsection{Policy Gradient}
In the field of LLM RL \citep{DBLP:conf/acl/TrungZJSJL24, openai_o1}, policy gradient-based (PG) algorithms \citep{DBLP:journals/ml/Williams92} are widely used.
Specifically, given an input prompt $\bm{x}$, an LLM $\pi_\theta$ sequentially generates a $T$-token response $\bm{y}=(y_1, ..., y_T)$: \begin{equation}
\pi_\theta(\bm{y}|\bm{x})=\textstyle \prod_{t=1}^T \pi_\theta(y_t|\bm{x}, \bm{y}_{<t}).
\end{equation}
Given a training dataset $\mathcal{D}=\{\bm{x}_1, ..., \bm{x}_N\}$ and reward function $R$, the RL objective is to maximize the expected reward:
\begin{equation}
J(\theta) = \mathbb{E}_{\bm{x}\sim\mathcal{D},\ \bm{y}\sim \pi_\theta(\cdot|\bm{x})} \left[R(\bm{x}, \bm{y})\right].
\end{equation}
PG algorithms then leverage gradient ascent to optimize the policy with the following gradient:
\begin{equation}
\nabla_\theta J(\theta) = \mathbb{E}_{\bm{x}\sim\mathcal{D},\ \bm{y}\sim \pi_\theta(\cdot|\bm{x})} \left[ \sum_{t=1}^{T} \nabla_\theta \log \pi_\theta(y_t|\bm{x}, \bm{y}_{<t}) \cdot A_t \right],
\end{equation}
where $A_t$ denotes the estimated advantage at time step $t$, i.e., how much better action $y_t$ is than the expected action under the current policy.

\subsection{Importance Sampling and PPO Objective}
To improve sample efficiency and adapt to modern infrastructure, mainstream RL algorithms for LLMs typically adopt a PPO-like surrogate objective \citep{DBLP:journals/corr/SchulmanWDRK17}:
\begin{equation}
J^{\textnormal{PPO}}(\theta) = \mathbb{E}_{\bm{x}\sim\mathcal{D},\ \bm{y}\sim \pi_{\theta_\textnormal{rollout}}(\cdot|\bm{x})}\sum_{t=1}^{T} \left[ \min(r_t\cdot A_t, \textnormal{clip}(r_t, 1-\varepsilon, 1+\varepsilon )\cdot A_t)\right],
\end{equation}
where $r_t = \frac{\pi_\theta(y_t|\bm{x},\bm{y}_{<t})}{\pi_{\theta_\textnormal{rollout}}(y_t|\bm{x},\bm{y}_{<t})}$ is the importance weight that corrects for the distribution mismatch, estimating the expected advantage of tokens generated by the behavior policy $\pi_{\theta_\textnormal{rollout}}$ under the target policy $\pi_\theta$. The clipping mechanism in PPO serves to implicitly enforce a trust region between the behavior and target policies, preventing overly large policy updates that could destabilize training.
The hyperparameter $\varepsilon \in (0,1)$ determines the width of this clipping interval.

We then analyze data with positive and negative advantages respectively. The policy gradient can then be expressed as:
\begin{equation}
    \nabla J^{\textnormal{PPO}} = \underbrace{\sum_{A_t>0} \pi_{\theta}(y_t) \cdot \mathbb{I}\{r_t<1+\varepsilon\} \cdot A_t \cdot \nabla \log \pi_\theta(y_t)}_{\textnormal{positive tokens}} + \underbrace{\sum_{A_t<0} \pi_{\theta}(y_t) \cdot \mathbb{I}\{r_t>1-\varepsilon\} \cdot A_t \cdot \nabla \log \pi_\theta(y_t)}_{\textnormal{negative tokens}},
    \label{equation:positive and negative samples}
\end{equation}
where $\mathbb{I}$ represents the indicator function.

\section{Motivation: Imbalanced Optimization and Entropy-Clip Rule}\label{sec:analysis}

In this section, we first conduct preliminary experiments to show the influence of data staleness on the RL optimization process. Next, we perform in-depth empirical and theoretical analysis to reveal the underlying mechanisms and provide new insights.

\begin{figure}[htbp]
\centering
\begin{minipage}[t]{0.48\textwidth}
\centering
\includegraphics[width=\linewidth]{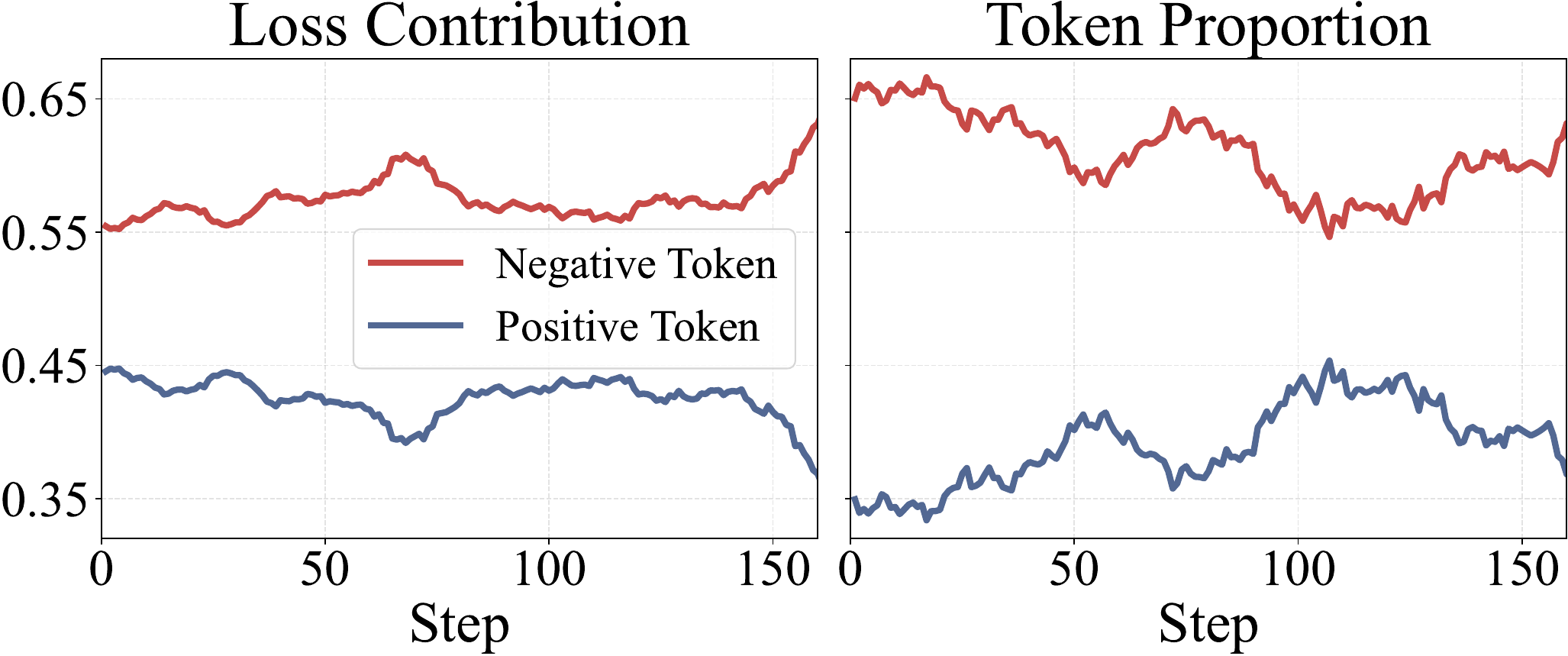} 
\vspace{-10pt}
\caption{Contribution of positive and negative tokens to the policy-gradient loss and their proportion of tokens during training.}
\label{fig:pos_contribution}
\end{minipage}
\hspace{0.01\textwidth}
\begin{minipage}[t]{0.48\textwidth}
\centering
\includegraphics[width=\linewidth]{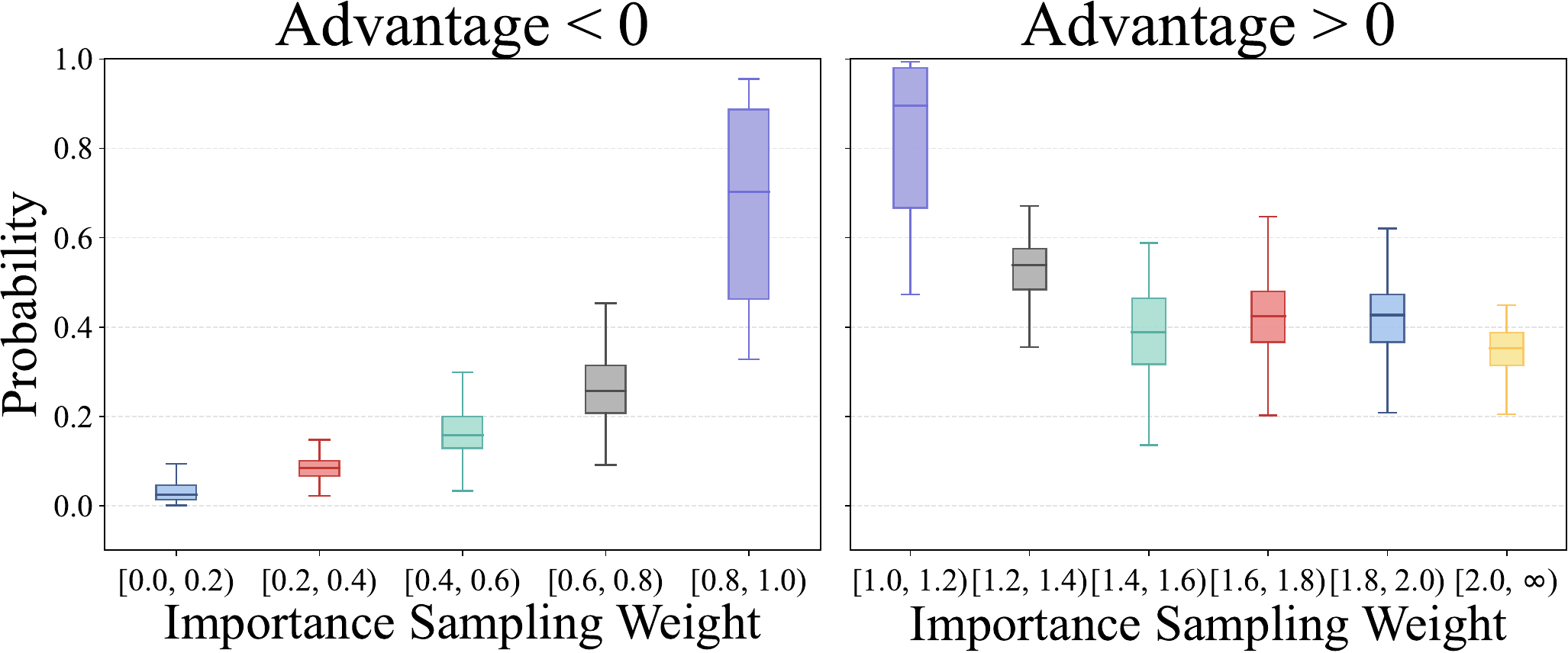} 
\vspace{-10pt}
\caption{Relationship between token probability and importance sampling weight.}
\label{fig:neg_pos_is_prob}
\vspace{-10pt}
\end{minipage}
\end{figure}

\paragraph{Training instability with data staleness.}
We perform experiments under different levels of data staleness using the popular GRPO algorithm. Results in Figure~\ref{fig:on_off_policy} show that, compared to on-policy training, off-policy RL typically suffers from instability, and entropy decreases rapidly, reflecting reduced exploratory capacity \citep{skywork_or1}. As staleness increases, the entropy decline becomes more severe and a larger number of tokens are clipped; meanwhile, training becomes more unstable. 
In the following paragraphs, we attempt to explain this phenomenon from different perspectives and summarize the motivation behind our method.

\paragraph{Excessive negative samples lead to imbalanced optimization.}

\begin{wrapfigure}{r}{0.27\linewidth} 
    \centering
    \includegraphics[width=\linewidth]{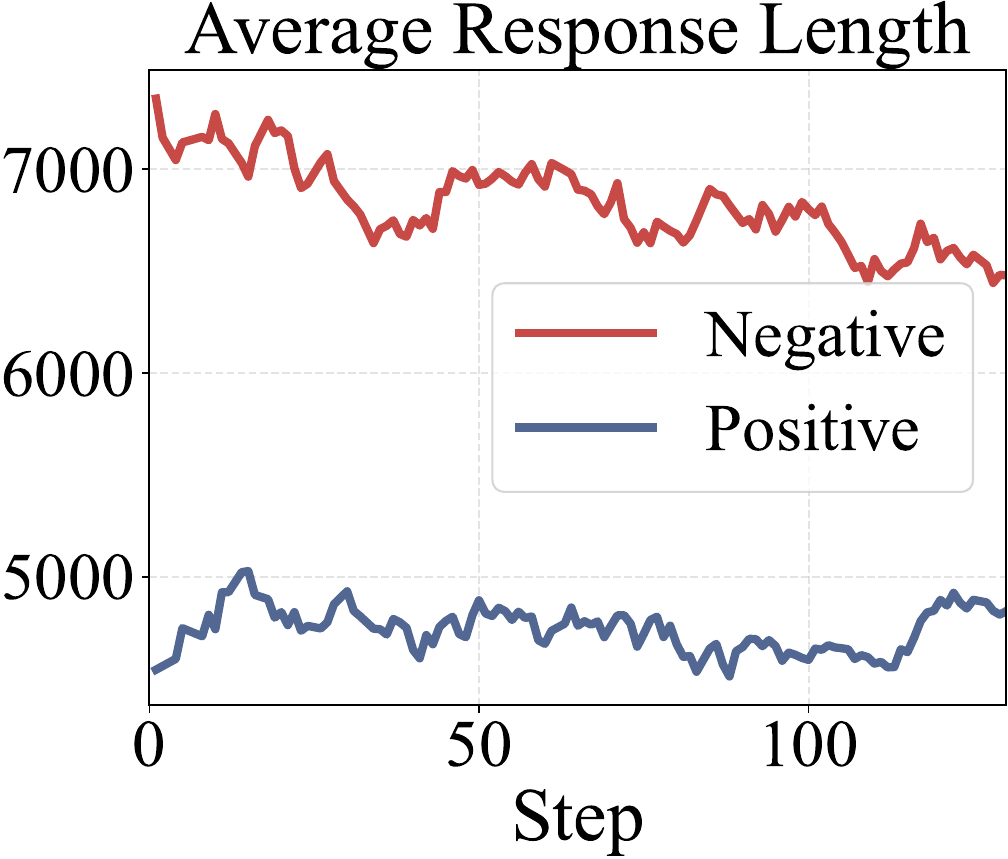}
    \vspace{-18pt}
    \caption{Average model response length during training.}
    \label{fig:average_resp_len}
\vspace{-18pt}
\end{wrapfigure}

Within the PPO-like objective for policy updates, we analyze tokens with positive and negative advantages separately, as shown in Equation \ref{equation:positive and negative samples}. Empirical results in Figure \ref{fig:pos_contribution} reveal a pronounced imbalance: positive samples constitute a minority both in number and in their contribution to the policy-gradient loss. We attribute this skew to two main factors: (i) the model tends to generate longer trajectories on difficult queries, thereby producing more tokens in negative samples (Figure \ref{fig:average_resp_len}); and (ii) in early stages of training, the model has not yet acquired sufficient capability, resulting in a higher proportion of negative samples. This observation may help explain the effectiveness of certain curriculum-based approaches \citep{DBLP:conf/icml/XiCHJZHDLGWGSFZ24, DBLP:journals/corr/abs-2507-22607}.

In the RL training of LLMs, reinforcing positive samples is often more efficient for driving performance gains than attempting to “suppress” the vast number of negative samples \citep{DBLP:journals/corr/abs-2308-08998, zhu2025flowrl}. To this end, prior work has proposed amplifying positive signals through the clip-higher technique \citep{DBLP:journals/corr/abs-2503-14476}. However, merely enlarging the clipping upper bound does not mitigate the influence of negative data, thus failing to prevent them from dominating the optimization process. Moreover, as shown in Equation \ref{equation:positive and negative samples}, the accumulation of low-probability negative tokens (i.e., $\pi_\theta(y_t) \to 0$, driving the log term toward $-\infty$) may trigger gradient explosion, further destabilizing training \citep{DBLP:journals/corr/abs-2505-12929}.

\paragraph{The Entropy-Clip Rule exposes insufficient entropy promotion in optimization, leading to entropy collapse.}

Theoretically, we derives Equation \ref{equation:entropy} (see Appendix~\ref{appendix:proof} for detailed derivations) for PPO surrogate objective to reveal the factors that influence the policy entropy \citep{DBLP:journals/corr/abs-2503-14286}: 

\begin{equation}
    \Delta\mathcal{H}(\pi_\theta)\approx-\eta \cdot \text{Cov}_{\bm{y} \sim \pi_\theta(\cdot|\bm{x})} \left[\log \pi_\theta(y_t|\bm{x},\bm{y_{<t}}), A_t \cdot \mathcal{X}(y_t) +C \right],
    \label{equation:entropy}
\end{equation}

where $C$ is a constant, and 

\begin{equation}
\mathcal{X}(y_t) =
\begin{cases}
1, & \bm{\mathit{if}}\; A_t > 0 \ \&\  r_t< 1+\epsilon \\
& \bm{\mathit{or}}\; A_t < 0 \ \&\ r_t>1-\epsilon \\
0, & \text{otherwise.}
\end{cases} 
\end{equation}

We observe that changes in policy entropy are driven by the influence of unclipped tokens, which is determined by the covariance between their log probabilities and advantages. We term this as \textbf{the Entropy-Clip Rule}.
The left side of Figure \ref{fig:main} illustrates how the optimization of different tokens influences the probability distribution, thereby affecting entropy. The Entropy-Clip Rule theoretically explains the following statement: Specifically, updating the policy with positive high-probability tokens (high advantage, high probability) and negative low-probability tokens (low advantage, low probability) sharpens the distribution and consequently reduces entropy. Conversely, updating the policy with negative high-probability tokens and positive low-probability tokens smooths the distribution, resulting in an increase in entropy (detailed proofs are available in Appendix \ref{analysis}).

Empirically, our statistical analysis on token probabilities and their importance sampling (IS) weights further clarifies this dynamic. As shown in Figure \ref{fig:neg_pos_is_prob}, we find that tokens with either very high or very low IS weights tend to have low probabilities. However, in standard algorithms with symmetric clipping bounds (e.g., [0.8,1.2]), a majority of positive, low-probability tokens are prevented from contributing to the optimization. This systematic exclusion of entropy-increasing updates causes a continuous decline in entropy, ultimately crippling the model's exploratory capacity and resulting in a performance bottleneck.

\begin{tcolorbox}[
colback=gblue!25, 
colframe=gblue!95, 
]
\textbf{Summary of motivation.} Based on the above analysis, we summarize two main motivations: (1) to balance the contributions of positive and negative tokens while preventing gradient explosion, and (2) to preserve policy entropy for sustaining exploration and preventing collapse.
\end{tcolorbox}

\section{Methodology}\label{sec:methods}

\subsection{Validation Experiment: Asymmetric Clipping}\label{sec:asymmetric}

\begin{wrapfigure}{r}{0.5\linewidth} 
\begin{minipage}{0.5\textwidth}
\centering
    \vspace{-12pt}
    \includegraphics[width=\linewidth]{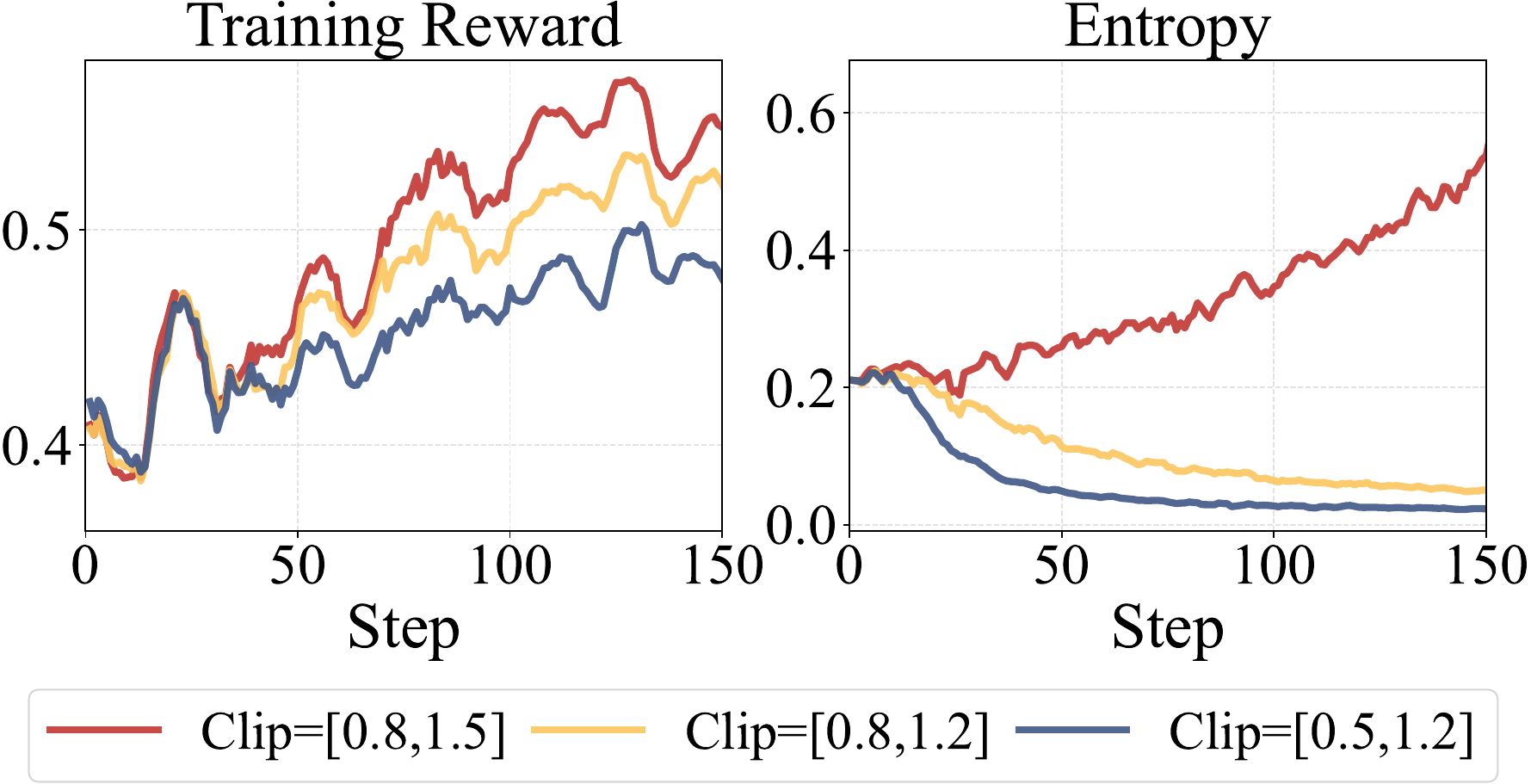}
\caption{Training dynamics of asymmetric clipping experiments.}
\label{fig:asymmetric_clip}
\end{minipage}
\vspace{-12pt}
\end{wrapfigure}

The main idea of our method is to stabilize the training and maintain exploration ability of the policy by asymmetrically adjusting the trust region for positive and negative tokens, i.e., adjusting $c_\textnormal{low}$ and $c_\textnormal{high}$.

We then conduct preliminary experiments to examine whether asymmetrically adjusting the clipping range could influence training dynamics. The results, shown in Figure \ref{fig:asymmetric_clip}, together with Figure \ref{fig:neg_pos_is_prob}, reveal that increasing the upper bound $c_\textnormal{high}$ (which introduces more low-probability positive tokens to policy updates) improves performance while counteracting the downward trend of entropy, albeit at a rapid pace. In contrast, relaxing the lower bound $c_\textnormal{low}$ (which introduces more low-probability negative tokens to policy updates) not only degrades performance but also accelerates entropy collapse. These findings confirm the effectiveness of entropy control through asymmetric clipping. Nevertheless, this strategy remains relatively rigid and manually specified, providing limited flexibility and adaptation.

\subsection{BAPO: BAlanced Policy Optimization with Adaptive Clipping}\label{sec:method}
To this end, we propose \textbf{BA}lanced \textbf{P}olicy \textbf{O}ptimization with Adaptive Clipping (\textbf{BAPO}), a new method to achieve stable, fast RL optimization for LLMs. The core insight of BAPO lies in its adaptive clipping mechanism, which dynamically adjusts the clipping bounds $c_\textnormal{high}$ and $c_\textnormal{low}$, to regulate the positive contribution to the policy loss and maintain a balance in entropy throughout RL training. Formally, for each update with a batch, our goal is to find a pair of $c_\textnormal{high}$ and $c_\textnormal{low}$ that satisfy:
\begin{equation}
    \frac{|\sum_{A_t>0} \pi_{\theta_{\textnormal{rollout}}}(y_t) \cdot \left[ \min(r_t\cdot A_t, \textnormal{clip}(r_t, 0, c_\textnormal{high} )\cdot A_t)\right]|}{|\sum_{A_t} \pi_{\theta_{\textnormal{rollout}}}(y_t) \cdot\left[ \min(r_t\cdot A_t, \textnormal{clip}(r_t, c_\textnormal{low}, c_\textnormal{high})\cdot A_t)\right]|} \geq \rho_0\ ,
\label{eqn:contribution_condition}
\end{equation}
where $\rho_0$ is the target contribution of positive signals to the policy gradient loss. 
Specifically, BAPO gradually increases $c_\textnormal{high}$ and $c_\textnormal{low}$ with step sizes of $\delta_1$ and $\delta_2$, respectively, until the condition in Equation \ref{eqn:contribution_condition} is met. We present an overview of BAPO in Figure \ref{fig:main} and summarize it in Algorithm \ref{algorithm}.

\begin{algorithm}[t]
  \SetAlgoLined
  \SetKwInOut{Input}{Input}\SetKwInOut{Output}{Output}
  \SetKwProg{myproc}{Procedure}{}{}
   \KwIn{Initialized LLM policy $\pi_{\theta}$, training dataset $\mathcal{D}$, reward function $R$, staleness $E$, movable range of clipping bounds $[a^-, b^-]$ and $[a^+, b^+]$, step size of upper bound $\delta_1$, step size of lower bound $\delta_2$, positive token contribution threshold $\rho_0$}
   \For{step $s=1...S$}{
   \myproc{\textnormal{Sample and filter out responses}}{
         Update the old LLM policy $\pi_{\theta_\textnormal{rollout}}\gets \pi_\theta$ ;\\
        Sample the $s$-th batch $\mathcal{D}_s$ from $\mathcal{D}$ ;\\
        Sample $G$ responses $\{\bm{y}_i\}_{i=1}^{G}\sim \pi_{\theta_\textnormal{rollout}}(\cdot|\bm{x})$, where $\bm{x}\in\mathcal{D}_s$ ; \\
        Compute reward and advantage for each $\bm{y}_i$ based on reward function $R$ ; \\
    }
    \For{staleness $=0...E$}{
    \myproc{\textnormal{Dynamically adjusting the clipping bounds} $c_\textnormal{high}$ and $c_\textnormal{low}$}{
     Initialize clipping bounds $c_\textnormal{low}=a^-$ and $c_\textnormal{high}=a^+$ ; \\
     \While{the positive token contribution $\rho < \rho_0$ \textnormal{\textbf{and}} $c_{\textnormal{low}}+\delta_2 \le b^-$ \\}{\eIf{$c_\textnormal{high}+\delta_1 \le b^+$}{$c_\textnormal{high}\gets c_\textnormal{high}+\delta_1$\\
      }{
      $c_\textnormal{low}\gets c_\textnormal{low}+\delta_2$\\
      }}
    }
     \myproc{\textnormal{Update the LLM policy} $\pi_\theta$}{
        Update the LLM policy $\pi_\theta$ by maximizing the following objective: \\
        \ \ \ \ $J^{\textnormal{BAPO}}(\theta) = \mathbb{E}_{\bm{y}\sim \pi_{\theta_\textnormal{rollout}}(\cdot|\bm{x})} \sum_{t=1}^{T} \left[ \min(r_t\cdot A_t, \textnormal{clip}(r_t, c_\textnormal{low}, c_\textnormal{high})\cdot A_t)\right]$
    }
    }
}
  \caption{BAPO}
  \label{algorithm}
\end{algorithm}

Overall, BAPO offers several significant benefits. First, by dynamically adjusting $c_\textnormal{high}$ and $c_\textnormal{low}$ for each step, we can increase the contribution of positive tokens to the policy-gradient loss while preventing negative tokens from excessively dominating the optimization objective. Second, based on our earlier analysis of the relationship between IS weights and token probabilities in Figure \ref{fig:neg_pos_is_prob}, BAPO incorporates more low-probability positive tokens and filters out more low-probability negative tokens, both of which contribute to maintaining entropy. Third, by setting the target contribution from positive tokens, BAPO prevents uncontrolled entropy growth, avoids situations where positive tokens overwhelm the loss, and mitigates tail degradation–where the model overfits to easy problems but fails to handle more challenging ones \citep{DBLP:conf/naacl/DingXHLZXCGZH25}.

\subsection{Analysis}

\paragraph{Stable and fast training of BAPO.} As shown in Figure \ref{fig:BAPO_training_dynamics}, BAPO enables a more stable optimization process, characterized by rapidly increasing training rewards, greater contributions from positive tokens, steady gradient normalization, and stable policy entropy–resulting in an improved balance between exploration and exploitation.

\begin{wrapfigure}{r}{0.28\linewidth}
\centering  
\vspace{-21pt}
\includegraphics[width=0.9\linewidth]{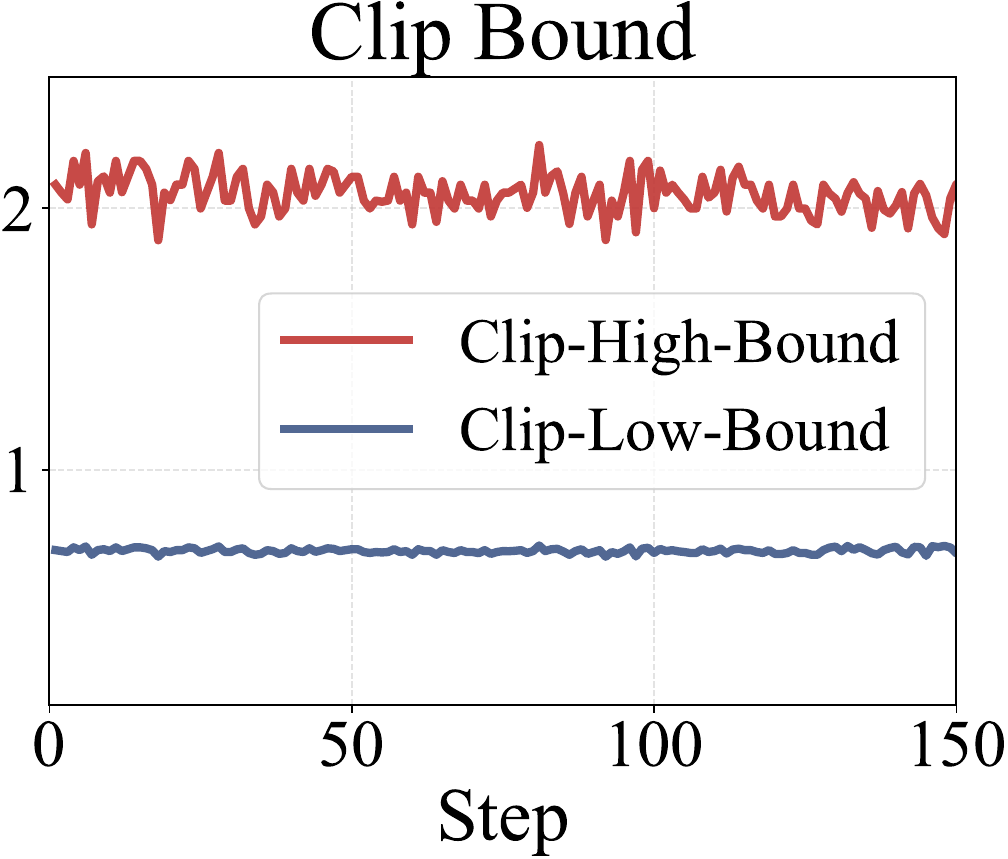}
\vspace{-8pt}
\caption{Clipping bounds.}
\label{fig:clip}
\vspace{-5pt}
\end{wrapfigure}

We further visualize the adjustment process of the clipping bounds in BAPO. As shown in Figure \ref{fig:clip}, the averaged upper and lower bounds both fluctuate during training, confirming that BAPO dynamically adjusts the clipping for both types of data and adaptively balances their contributions to the loss. In contrast to approaches such as DAPO \citep{DBLP:journals/corr/abs-2503-14476} or the asymmetric clipping in Section \ref{sec:asymmetric}, which rely on empirical tuning, BAPO eliminates the need for complex manual hyperparameter tuning, making it simple yet effective.

\begin{figure}[t]
\centering
    \includegraphics[width=0.9\linewidth]{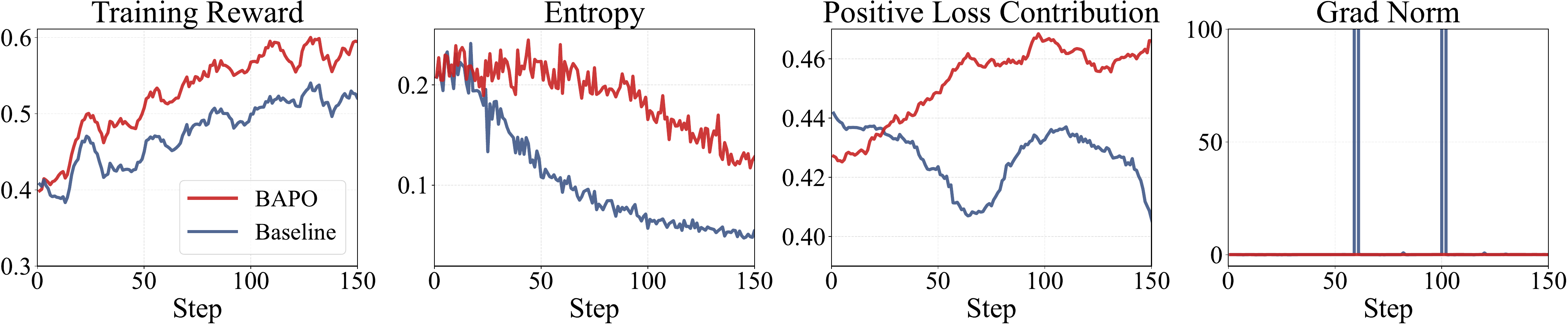}
\vspace{-1pt}
\caption{Training dynamics of BAPO.}
\label{fig:BAPO_training_dynamics}
\end{figure}

\begin{figure}[t]
\begin{center}
\includegraphics[width=0.9\linewidth]{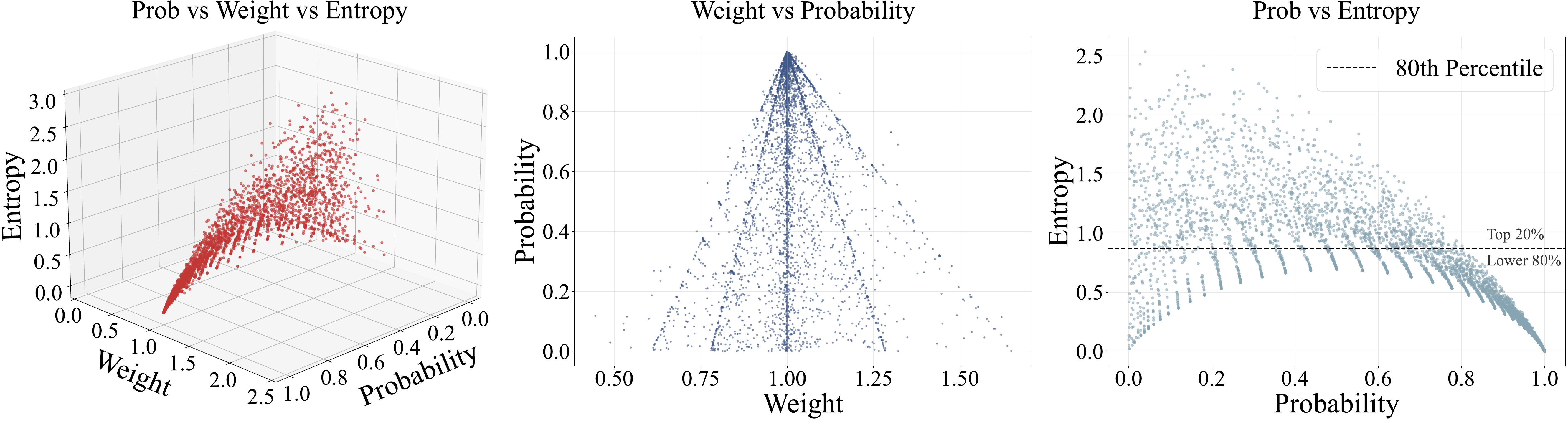}
\end{center}
\vspace{-1pt}
\caption{Relationship among token probabilities, importance sampling weights, and entropy.}
\label{fig:token_info}
\end{figure}

\begin{wrapfigure}{r}{0.5\linewidth}
\vspace{-18pt}
\begin{center}
\includegraphics[width=0.9\linewidth]{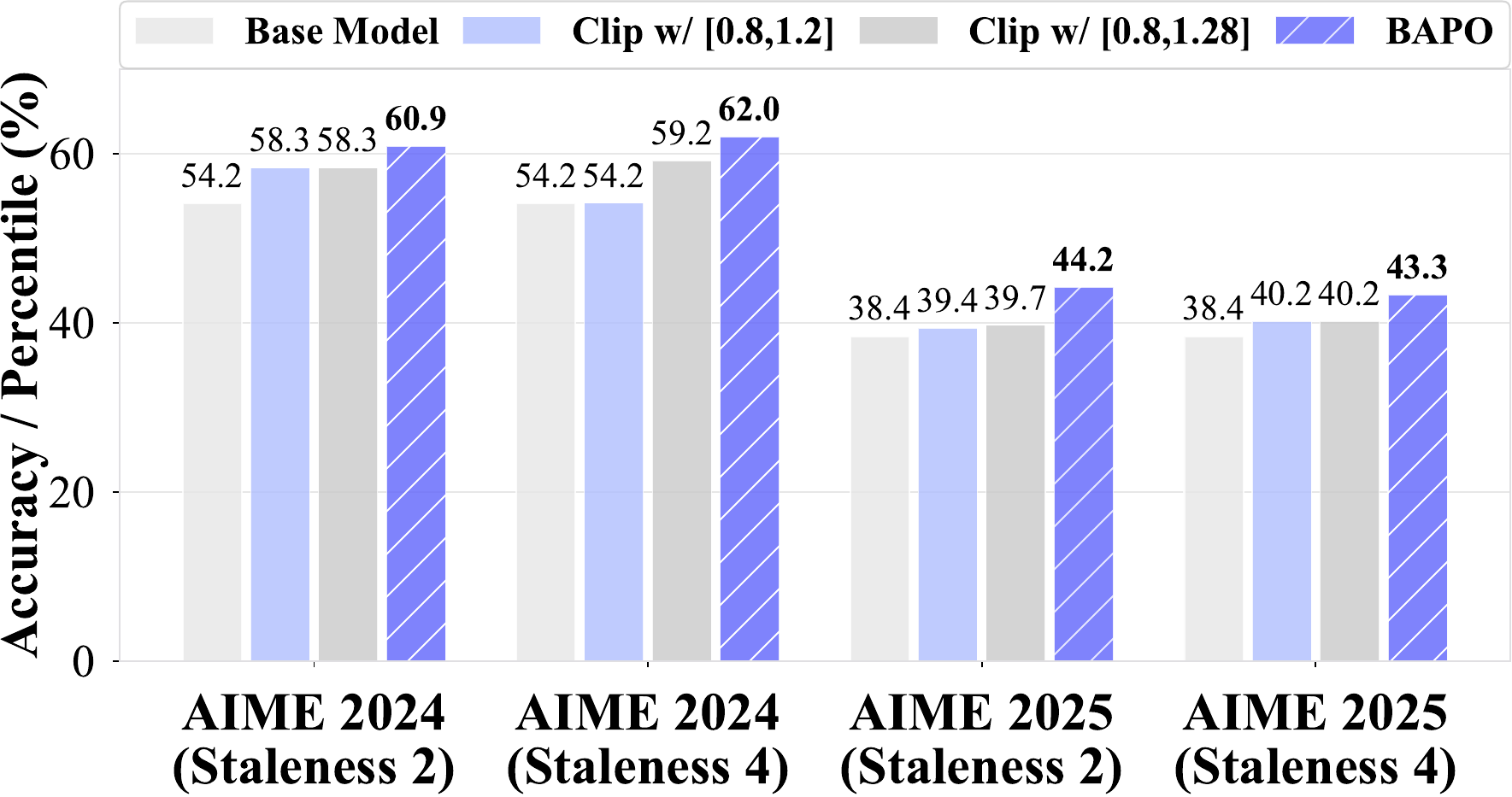}
\end{center}
\vspace{-15pt}
\caption{Results with different data staleness.}
\label{fig:results_r1_distill}
\vspace{-30pt}
\end{wrapfigure}

\paragraph{Effectiveness of BAPO across different staleness.}
We conduct experiments using the R1-Distill model \citep{DeepSeek-R1} on the SkyWork-OR1-RL dataset \citep{skywork_or1}, with a maximum sequence length of $32k$. The results in Figure \ref{fig:results_r1_distill} show that under different data staleness, our method consistently outperforms both the baseline and the clip-higher approach, demonstrating its superiority.

\paragraph{The working mechanism of BAPO and its connection to prior work.}
To better understand the working mechanism of BAPO, we present the relationship among token probabilities, IS weights, and entropy during training in Figure \ref{fig:token_info}. We find that as IS weights deviate further from $1$, the corresponding token probabilities decrease, and such low-probability tokens often exhibit higher entropy. Based on this observation, we explain how BAPO relates to prior work. For example, Clip-Higher in \citet{DBLP:journals/corr/abs-2503-14476} sets the clipping upper bound to $1.28$, thereby including more low-probability positive tokens in training, which stabilizes entropy while balancing the contributions of positive and negative tokens. Similarly, \citet{DBLP:journals/corr/abs-2506-01939} retain only the top 20\% highest-entropy tokens for training, ensuring stable entropy throughout optimization and preserving the model’s exploratory capability, and the target entropy technique in \citet{skywork_or1} plays a similar role, which aligns with our motivation.

\section{Experiments and Discussion}\label{sec:experiments}

\begin{table*}[t]
\centering
\caption{
Main evaluation results.
}
\resizebox{0.99\textwidth}{!}{ 
\begin{tabular}{lcccc}
\toprule
\textbf{Model} & \textbf{Model Size}  & \textbf{AIME 2024} & \textbf{AIME 2025} & \textbf{Average} \\
\midrule
\rowcolor{gray!10}\multicolumn{5}{c}{\emph{$\geq$ 100B Models and Proprietary Models }} \\
\texttt{Qwen3-235B-A22B} \citep{DBLP:journals/corr/abs-2505-09388} & 235B &  $85.7$ & $81.5$ & $83.6$ \\
\texttt{DeepSeek-R1} \citep{DeepSeek-R1}  & 671B &  $79.8$ & $70.0$ & $74.9$ \\
\texttt{DeepSeek-R1-0528} \citep{DeepSeek-R1} & 671B &  $91.4$ & $87.5$ & $89.5$ \\
\texttt{o1$_\texttt{medium}$} \citep{openai_o1} &  - & $83.3$ & $79.0$ & $81.2$ \\
\texttt{o3-mini$_\texttt{medium}$} \citep{openai_o3} &  - & $79.6$ & $76.7$ & $78.2$ \\
\texttt{o3-mini$_\texttt{high}$} \citep{openai_o3} &  - & $87.3$ & $86.5$ & $86.9$ \\
\texttt{Gemini-2.0$_\texttt{Flash-Thinking}$} \citep{GoogleGemini2024Update} &  - & $73.3$ & $53.5$ & $63.4$ \\
\texttt{Gemini-2.5$_\texttt{Flash-Thinking-0520}$} \citep{DBLP:journals/corr/abs-2507-06261} &  - & $82.3$ & $72.0$ & $77.2$ \\
\midrule
\rowcolor{gray!10}\multicolumn{5}{c}{\emph{10B - 100B Scale Models }} \\
\texttt{Qwen3-30B-A3B} \citep{DBLP:journals/corr/abs-2505-09388} & 30B & $-$ & $61.3$ & $-$ \\
\texttt{R1-Distill-Qwen-32B} \citep{DeepSeek-R1} & 32B & $72.6$ & $54.9$ & $63.8$ \\
\texttt{QwQ-32B} \citep{qwq} & 32B &  $79.5$ & $65.3$ & $72.4$\\
\texttt{Qwen3-32B} \citep{DBLP:journals/corr/abs-2505-09388} & 32B &  $81.4$ & $72.9$ & $77.2$ \\
\texttt{SkyWork-OR1-32B} \citep{skywork_or1} & 32B &  $82.2$ & $73.3$ & $77.8$ \\
\texttt{BP-Math-32B$_{\texttt{SFT}}$} & 32B &  $84.4$ & $78.1$ & $81.3$ \\
\texttt{BP-Math-32B$_{\texttt{GRPO}}$} & 32B &  $84.6$ & $78.8$ & $81.7$ \\
\texttt{BP-Math-32B$_{\texttt{BAPO}}$} & 32B & $\mathbf{87.1}$ & $\mathbf{80.0}$ & $\mathbf{83.5}$ \\
\midrule
\rowcolor{gray!10}\multicolumn{5}{c}{\emph{$\leq$ 10B Models}} \\
\texttt{R1-Distill-Qwen-7B} \citep{DeepSeek-R1} & 7B &  $54.2$ & $38.4$ & $46.3$ \\
\texttt{Light-R1-7B-DS} \citep{DBLP:journals/corr/abs-2503-10460} & 7B  & $59.1$ & $44.2$ & $51.7$ \\
\texttt{AReaL-boba-RL-7B} \citep{DBLP:journals/corr/abs-2505-24298} & 7B & $61.9$ & $48.3$ & $55.1$ \\
\texttt{AceReason-Nemotron-7B} \citep{DBLP:journals/corr/abs-2505-16400} & 7B & $69.0$ & $53.6$ & $61.3$ \\
\texttt{SkyWork-OR1-7B} \citep{skywork_or1} & 7B &  $70.2$ & $54.6$ & $62.4$ \\
\texttt{BP-Math-7B}$_{\texttt{SFT}}$ & 7B &  $66.9$ & $59.0$ & $62.9$ \\
\texttt{BP-Math-7B}$_{\texttt{GRPO}}$ & 7B &  $69.2$ & $59.2$ & $64.2$ \\
\texttt{BP-Math-7B}$_{\texttt{BAPO}}$ & 7B &  $\mathbf{70.8}$ & $\mathbf{62.5}$ & $\mathbf{66.7}$ \\
\bottomrule
\end{tabular}
}
\label{tab:main}
\end{table*}

\subsection{Experimental Setups}
\paragraph{Datasets and Models.}
We use SkyWork-OR1-RL-Data \citep{skywork_or1} as our RL dataset, as it is widely adopted and of high quality. For evaluation, we employ both the AIME 2024 and the newly released AIME 2025 \citep{aime} benchmarks. Our experiments cover a range of backbone models, including \texttt{DeepSeek-R1-Distill-Qwen-7B}, \texttt{DeepSeek-R1-Distill-Qwen-32B} \citep{DeepSeek-R1}, and \texttt{OctoThinker-Llama3.2-3B-Long-Zero} \citep{DBLP:journals/corr/abs-2506-20512}. In addition, we incorporate two our own supervised fine-tuning (SFT) models, \texttt{BP-Math-7B} and \texttt{BP-Math-32B}, which are derived from \texttt{Qwen2.5-Math} \citep{DBLP:journals/corr/abs-2409-12122} through fine-tuning.

\paragraph{Implementation details.} 
We leverage GRPO as the basis for BAPO.
Both our preliminary and validation experiments are conducted using \texttt{DeepSeek-R1-Distill-Qwen-7B}, with the maximum response length set to $8k$, learning rate to $2 \times 10^{-6}$, and temperature to $0.6$. For main results on BP-Math models, we set the maximum response length to $64k$ to align with the SFT setting. To introduce staleness, we adopt multiple strategies, including experience reuse through ppo\_epoch \citep{DBLP:journals/corr/SchulmanWDRK17} and the modern partial rollouts \citep{DBLP:journals/corr/abs-2501-12599, DBLP:journals/corr/abs-2505-24298}. For BAPO, we set the target contribution $\rho_0=0.4$, the movable range $a^{-}=0.6$, $b^{-}=0.9$, $a^{+}=1.2$, $b^{+}=3.0$, and the step size $\delta_1=0.05$, $\delta_2=0.02$. These hyperparameters are not finely tuned, as they already demonstrate strong empirical performance. For evaluation, we report results averaged over $16$ rollouts.

\paragraph{Baselines.}
We include a variety of commercial and open-source models of different scales as baselines, as shown in Table \ref{tab:main}, and report their performance as extracted from prior work. In addition, we compare different training approaches, including SFT and GRPO.

\subsection{Main Results}
The main results are shown in Figure \ref{fig:results_r1} and Table \ref{tab:main}.

\paragraph{Significant performance improvements across models of varying sizes.}
For strong SFT models, GRPO provides only marginal benefits–for instance, it improves performance by just $0.2$ and $0.7$ points on AIME24 and AIME25 with the \texttt{BP-Math-32B} model. In contrast, BAPO delivers substantial gains across models of different scales. Specifically, with the \texttt{BP-Math-32B} model, BAPO outperforms SFT by $2.7$ and $1.9$ points on AIME24 and AIME25, respectively; with the \texttt{BP-Math-7B} model, it achieves even larger improvements of $3.9$ and $3.5$ points.

\paragraph{SOTA performance over open-source models of comparable sizes and competitive results against proprietary models.}
Compared to open-source models of similar sizes, our BAPO-trained models achieve state-of-the-art (SOTA) performance. For instance, among 32B models, \texttt{BP-Math-32B}$_\texttt{BAPO}$ outperforms \texttt{Qwen3-32B} by $5.7$ and $7.1$ points on AIME24 and AIME25, respectively, and surpasses \texttt{SkyWork-OR1-32B} by $4.9$ and $6.7$ points. Among 7B models, \texttt{BP-Math-7B}$_\texttt{BAPO}$ also delivers a notable $7.9$-point improvement over \texttt{SkyWork-OR1-7B} on AIME25.

Moreover, \texttt{BP-Math-32B}$_\texttt{BAPO}$ even outperforms some larger-scale models–for example, it surpasses \texttt{DeepSeek-R1} by $7.3$ and $10.0$ points on AIME24 and AIME25, respectively–while achieving performance comparable to \texttt{o3-mini}. Notably, even the smaller \texttt{BP-Math-7B}$_\texttt{BAPO}$ yields results on par with \texttt{Gemini-2.0-Flash-Thinking}, underscoring the competitiveness of our approach against commercial models.

\subsection{Discussion}\label{sec:results_llama}
\paragraph{Partial rollout.}
\begin{wrapfigure}{r}{0.5\linewidth} 
\begin{minipage}{0.5\textwidth}
    \vspace{-13pt}
    \centering
    \includegraphics[width=0.9\linewidth]{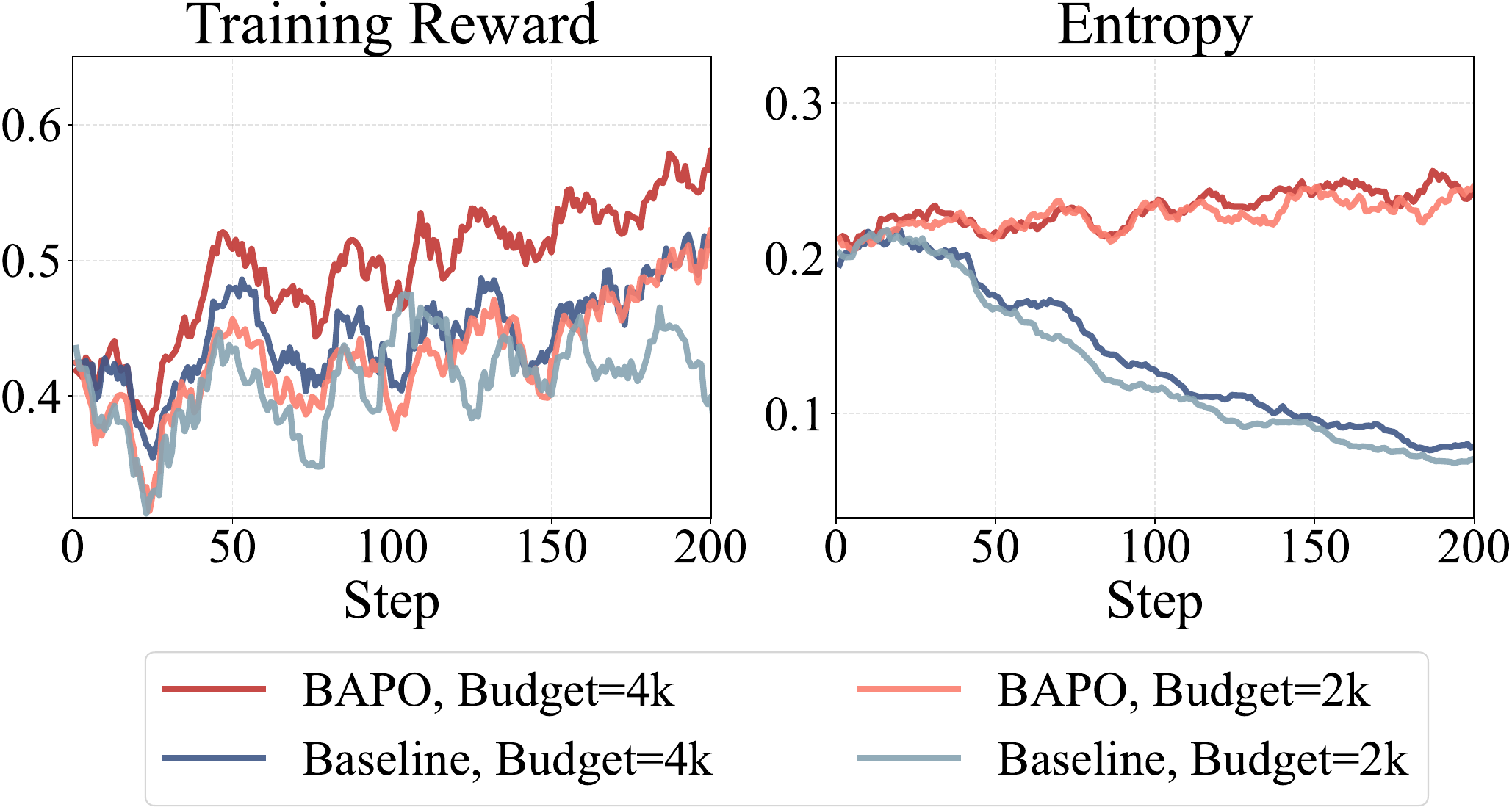}
    \vspace{-5pt}
    \caption{Training dynamics with partial rollout.}
    \label{fig:partial_rollout}
\end{minipage}
\end{wrapfigure}
To speed up rollouts in LLM reinforcement learning, modern AI infrastructures have introduced several techniques, with partial rollout being particularly noteworthy \citep{DBLP:journals/corr/abs-2501-12599, DBLP:journals/corr/abs-2505-24298}. In this approach, long trajectories are split into segments: when a rollout exceeds a fixed token budget, the unfinished portion is stored in a replay buffer and resumed in later iterations instead of being regenerated from scratch. While this improves training efficiency, it also introduces off-policy learning, since different parts of the same trajectory may come from multiple outdated policies.
We evaluate BAPO under this setting, as shown in Figure \ref{fig:partial_rollout}. Compared to the baseline GRPO, BAPO exhibits greater robustness to such off-policy infrastructures and achieves more stable optimization.

\paragraph{Results on OctoThinker-Llama3.2-3B-Long-Zero.}

\begin{wraptable}{r}{0.4\textwidth}
\vspace{-14pt}
    \caption{Performance of Llama-based models.}
    \vspace{-5pt}
    \label{tab:llama_result}
    \centering
    \resizebox{0.98\linewidth}{!}{
    \begin{tabular}{cccc}
    \toprule
    \textbf{Method} & \textbf{AIME 2024} & \textbf{AIME 2025} & \textbf{MATH} \\
    \midrule
    \texttt{GRPO}   & $2.5\%$ & $2.9\%$ & $58.4\%$ \\
    \midrule
    \texttt{BAPO}   & $5.4\%$ & $5.8\%$ & $66.0\%$ \\
    \bottomrule
    \end{tabular}
    }
\end{wraptable}

In addition to the DeepSeek-R1-Distill-Qwen, we also conducted experiments on Llama-based models \citep{DBLP:journals/corr/abs-2506-20512}. As shown in Table~\ref{tab:llama_result} and Figure \ref{fig:Llama_training_dynamics} in Appendix~\ref{appendix:llama}, our method achieves more competitive results and exhibits greater stability in training dynamics.

\section{Related Work}

Recent landmark models, like OpenAI o1 \citep{openai_o1}, DeepSeek-R1 \citep{DeepSeek-R1}, Gemini 2.5 \citep{DBLP:journals/corr/abs-2507-06261}, QwQ \citep{qwq}, have demonstrated that reinforcement learning can effectively enable long chain-of-thought reasoning in LLMs \citep{DBLP:journals/corr/abs-2402-03300e, zhang2025survey}. Mainstream algorithms include PPO \citep{DBLP:journals/corr/SchulmanWDRK17} and GRPO \citep{DBLP:journals/corr/abs-2402-03300e}: PPO constrains updates via a clipping-based surrogate objective, while GRPO enhances long-horizon reasoning through group-based rewards. 

Despite the remarkable success of RL for LLMs, ensuring stability and efficiency in optimization remains a major challenge \citep{DBLP:journals/corr/abs-2503-14476, DBLP:journals/corr/abs-2505-22617}. Recent studies have sought to better understand the underlying mechanisms of RL and proposed new methods to achieve a balance \citep{DBLP:journals/corr/abs-2505-22617, DBLP:journals/corr/abs-2507-18071, DBLP:journals/corr/abs-2506-01939, yang2025dcpo}. For example, DAPO \citep{DBLP:journals/corr/abs-2503-14476} introduces techniques such as Clip-Higher and dynamic sampling to raise the performance ceiling; \citet{DBLP:journals/corr/abs-2506-01939} explore optimizing only a small subset of high-entropy tokens for improved efficiency. \citet{skywork_or1}, \citet{DBLP:journals/corr/abs-2505-22617}, and other works \citep{DBLP:journals/corr/abs-2507-18071, DBLP:journals/corr/abs-2506-14758, DBLP:journals/corr/abs-2503-20783} systematically investigate how to maintain entropy stability during training, thereby preserving the model’s exploration ability. For off-policy RL, \citet{DBLP:journals/corr/abs-2503-14286} and \citet{DBLP:journals/corr/abs-2506-20520} introduce asymmetric clipping mechanisms. The most similar to our work is DCPO \citep{yang2025dcpo}, which adjusts token-level clipping based on token prior probabilities. However, our approach takes a holistic optimization perspective: we observe the imbalance in loss contributions and derive the Entropy-Clip Rule for the PPO objective, enabling dynamic control over global clipping bounds. We further validate the effectiveness of our method through larger-scale experiments.

\section{Conclusion}
In this paper, we begin by analyzing the impact of data staleness on model training through both empirical and theoretical studies. We reveal the imbalance between positive and negative samples in RL optimization, and derive as well as empirically validate the Entropy-Clip Rule for PPO-like objectives. Building on these insights, we propose BAPO, which dynamically adjusts the clipping bounds to balance positive and negative samples while preserving the model’s exploratory capability during training. We conduct extensive experiments across different models and settings to validate our method. We hope our work provides key insights for the LLM RL community.

\bibliography{main}

\begin{thebibliography}{38}
\providecommand{\natexlab}[1]{#1}
\providecommand{\url}[1]{\texttt{#1}}
\expandafter\ifx\csname urlstyle\endcsname\relax
  \providecommand{\doi}[1]{doi: #1}\else
  \providecommand{\doi}{doi: \begingroup \urlstyle{rm}\Url}\fi

\bibitem[AIME(2025)]{aime}
AIME.
\newblock Aime problems and solution, 2025.
\newblock URL \url{https://artofproblemsolving.com/wiki/index.php/AIME_Problems_and_Solutions}.

\bibitem[Anthropic(2025)]{claude_code}
Anthropic.
\newblock Claude code, 2025.
\newblock URL \url{https://docs.anthropic.com/en/docs/claude-code}.

\bibitem[Arnal et~al.(2025)Arnal, Narozniak, Cabannes, Tang, Kempe, and Munos]{DBLP:journals/corr/abs-2506-20520}
Charles Arnal, Ga{\"{e}}tan Narozniak, Vivien Cabannes, Yunhao Tang, Julia Kempe, and R{\'{e}}mi Munos.
\newblock Asymmetric {REINFORCE} for off-policy reinforcement learning: Balancing positive and negative rewards.
\newblock \emph{CoRR}, abs/2506.20520, 2025.
\newblock \doi{10.48550/ARXIV.2506.20520}.
\newblock URL \url{https://doi.org/10.48550/arXiv.2506.20520}.

\bibitem[Bai et~al.(2025)Bai, Bao, Chen, Chen, Chen, Chen, Chen, Chen, Chen, Chen, Cui, Ding, Dong, Du, Du, Du, Du, Fan, Feng, Fu, Gao, Gao, Gao, Gao, Gu, Guan, Guo, Guo, Hu, Hao, He, He, He, Hong, Hu, Hu, Huang, Huang, Huang, Jiang, Jiang, Jin, Kang, Lai, Li, Li, Li, Li, Li, Li, Li, Li, Li, Lin, Lin, Lin, Liu, Liu, Liu, Liu, Liu, Liu, Liu, Liu, Liu, Liu, Liu, Liu, Liu, Liu, Liu, Lu, Lu, Ma, Ma, Ma, Mao, Mei, Men, Miao, Pan, Peng, Qin, Qu, Shang, Shi, Shi, Song, Su, Su, Sun, Sung, Tang, Tao, Teng, Wang, Wang, Wang, and Wang]{DBLP:journals/corr/abs-2507-20534}
Yifan Bai, Yiping Bao, Guanduo Chen, Jiahao Chen, Ningxin Chen, Ruijue Chen, Yanru Chen, Yuankun Chen, Yutian Chen, Zhuofu Chen, Jialei Cui, Hao Ding, Mengnan Dong, Angang Du, Chenzhuang Du, Dikang Du, Yulun Du, Yu~Fan, Yichen Feng, Kelin Fu, Bofei Gao, Hongcheng Gao, Peizhong Gao, Tong Gao, Xinran Gu, Longyu Guan, Haiqing Guo, Jianhang Guo, Hao Hu, Xiaoru Hao, Tianhong He, Weiran He, Wenyang He, Chao Hong, Yangyang Hu, Zhenxing Hu, Weixiao Huang, Zhiqi Huang, Zihao Huang, Tao Jiang, Zhejun Jiang, Xinyi Jin, Yongsheng Kang, Guokun Lai, Cheng Li, Fang Li, Haoyang Li, Ming Li, Wentao Li, Yanhao Li, Yiwei Li, Zhaowei Li, Zheming Li, Hongzhan Lin, Xiaohan Lin, Zongyu Lin, Chengyin Liu, Chenyu Liu, Hongzhang Liu, Jingyuan Liu, Junqi Liu, Liang Liu, Shaowei Liu, T.~Y. Liu, Tianwei Liu, Weizhou Liu, Yangyang Liu, Yibo Liu, Yiping Liu, Yue Liu, Zhengying Liu, Enzhe Lu, Lijun Lu, Shengling Ma, Xinyu Ma, Yingwei Ma, Shaoguang Mao, Jie Mei, Xin Men, Yibo Miao, Siyuan Pan, Yebo Peng, Ruoyu Qin, Bowen Qu, Zeyu Shang,
  Lidong Shi, Shengyuan Shi, Feifan Song, Jianlin Su, Zhengyuan Su, Xinjie Sun, Flood Sung, Heyi Tang, Jiawen Tao, Qifeng Teng, Chensi Wang, Dinglu Wang, Feng Wang, and Haiming Wang.
\newblock Kimi {K2:} open agentic intelligence.
\newblock \emph{CoRR}, abs/2507.20534, 2025.
\newblock \doi{10.48550/ARXIV.2507.20534}.
\newblock URL \url{https://doi.org/10.48550/arXiv.2507.20534}.

\bibitem[Chen et~al.(2025)Chen, Yang, Liu, Lee, Xu, Shoeybi, Catanzaro, and Ping]{DBLP:journals/corr/abs-2505-16400}
Yang Chen, Zhuolin Yang, Zihan Liu, Chankyu Lee, Peng Xu, Mohammad Shoeybi, Bryan Catanzaro, and Wei Ping.
\newblock Acereason-nemotron: Advancing math and code reasoning through reinforcement learning.
\newblock \emph{CoRR}, abs/2505.16400, 2025.
\newblock \doi{10.48550/ARXIV.2505.16400}.
\newblock URL \url{https://doi.org/10.48550/arXiv.2505.16400}.

\bibitem[Cheng et~al.(2025)Cheng, Huang, Zhu, Dai, Zhao, Zhang, and Wei]{DBLP:journals/corr/abs-2506-14758}
Daixuan Cheng, Shaohan Huang, Xuekai Zhu, Bo~Dai, Wayne~Xin Zhao, Zhenliang Zhang, and Furu Wei.
\newblock Reasoning with exploration: An entropy perspective.
\newblock \emph{CoRR}, abs/2506.14758, 2025.
\newblock \doi{10.48550/ARXIV.2506.14758}.
\newblock URL \url{https://doi.org/10.48550/arXiv.2506.14758}.

\bibitem[Comanici et~al.(2025)Comanici, Bieber, Schaekermann, Pasupat, Sachdeva, Dhillon, Blistein, Ram, Zhang, Rosen, Marris, Petulla, Gaffney, Aharoni, Lintz, Pais, Jacobsson, Szpektor, Jiang, Haridasan, Omran, Saunshi, Bahri, Mishra, Chu, Boyd, Hekman, Parisi, Zhang, Kawintiranon, Bedrax{-}Weiss, Wang, Xu, Purkiss, Mendlovic, Deutel, Nguyen, Langley, Korn, Rossazza, Ram{\'{e}}, Waghmare, Miller, Byrd, Sheshan, Bhardwaj, Janus, Rissa, Horgan, Silver, Wahid, Brin, Raimond, Kloboves, Wang, Gundavarapu, Shumailov, Wang, Pajarskas, Heyward, Nikoltchev, Kula, Zhou, Garrett, Kafle, Arik, Goel, Yang, Park, Kojima, Mahmoudieh, Kavukcuoglu, Chen, Fritz, Bulyenov, Roy, Paparas, Shemtov, Chen, Strudel, Reitter, Roy, Vlasov, Ryu, Leichner, Yang, Mariet, Vnukov, Sohn, Stuart, Liang, Chen, Rawlani, Koh, Co{-}Reyes, Lai, Banzal, Vytiniotis, Mei, and Cai]{DBLP:journals/corr/abs-2507-06261}
Gheorghe Comanici, Eric Bieber, Mike Schaekermann, Ice Pasupat, Noveen Sachdeva, Inderjit~S. Dhillon, Marcel Blistein, Ori Ram, Dan Zhang, Evan Rosen, Luke Marris, Sam Petulla, Colin Gaffney, Asaf Aharoni, Nathan Lintz, Tiago~Cardal Pais, Henrik Jacobsson, Idan Szpektor, Nan{-}Jiang Jiang, Krishna Haridasan, Ahmed Omran, Nikunj Saunshi, Dara Bahri, Gaurav Mishra, Eric Chu, Toby Boyd, Brad Hekman, Aaron Parisi, Chaoyi Zhang, Kornraphop Kawintiranon, Tania Bedrax{-}Weiss, Oliver Wang, Ya~Xu, Ollie Purkiss, Uri Mendlovic, Ila{\"{\i}} Deutel, Nam Nguyen, Adam Langley, Flip Korn, Lucia Rossazza, Alexandre Ram{\'{e}}, Sagar Waghmare, Helen Miller, Nathan Byrd, Ashrith Sheshan, Raia Hadsell~Sangnie Bhardwaj, Pawel Janus, Tero Rissa, Dan Horgan, Sharon Silver, Ayzaan Wahid, Sergey Brin, Yves Raimond, Klemen Kloboves, Cindy Wang, Nitesh~Bharadwaj Gundavarapu, Ilia Shumailov, Bo~Wang, Mantas Pajarskas, Joe Heyward, Martin Nikoltchev, Maciej Kula, Hao Zhou, Zachary Garrett, Sushant Kafle, Sercan Arik, Ankita Goel,
  Mingyao Yang, Jiho Park, Koji Kojima, Parsa Mahmoudieh, Koray Kavukcuoglu, Grace Chen, Doug Fritz, Anton Bulyenov, Sudeshna Roy, Dimitris Paparas, Hadar Shemtov, Bo{-}Juen Chen, Robin Strudel, David Reitter, Aurko Roy, Andrey Vlasov, Changwan Ryu, Chas Leichner, Haichuan Yang, Zelda Mariet, Denis Vnukov, Tim Sohn, Amy Stuart, Wei Liang, Minmin Chen, Praynaa Rawlani, Christy Koh, JD~Co{-}Reyes, Guangda Lai, Praseem Banzal, Dimitrios Vytiniotis, Jieru Mei, and Mu~Cai.
\newblock Gemini 2.5: Pushing the frontier with advanced reasoning, multimodality, long context, and next generation agentic capabilities.
\newblock \emph{CoRR}, abs/2507.06261, 2025.
\newblock \doi{10.48550/ARXIV.2507.06261}.
\newblock URL \url{https://doi.org/10.48550/arXiv.2507.06261}.

\bibitem[Cui et~al.(2025)Cui, Zhang, Chen, Yuan, Wang, Zuo, Li, Fan, Chen, Chen, Liu, Peng, Bai, Ouyang, Cheng, Zhou, and Ding]{DBLP:journals/corr/abs-2505-22617}
Ganqu Cui, Yuchen Zhang, Jiacheng Chen, Lifan Yuan, Zhi Wang, Yuxin Zuo, Haozhan Li, Yuchen Fan, Huayu Chen, Weize Chen, Zhiyuan Liu, Hao Peng, Lei Bai, Wanli Ouyang, Yu~Cheng, Bowen Zhou, and Ning Ding.
\newblock The entropy mechanism of reinforcement learning for reasoning language models.
\newblock \emph{CoRR}, abs/2505.22617, 2025.
\newblock \doi{10.48550/ARXIV.2505.22617}.
\newblock URL \url{https://doi.org/10.48550/arXiv.2505.22617}.

\bibitem[Ding et~al.(2025)Ding, Xi, He, Lizhuoyuan, Zhai, Xiaowei, Cai, Gui, Zhang, and Huang]{DBLP:conf/naacl/DingXHLZXCGZH25}
Yiwen Ding, Zhiheng Xi, Wei He, Lizhuoyuan Lizhuoyuan, Yitao Zhai, Shi Xiaowei, Xunliang Cai, Tao Gui, Qi~Zhang, and Xuanjing Huang.
\newblock Mitigating tail narrowing in {LLM} self-improvement via socratic-guided sampling.
\newblock In Luis Chiruzzo, Alan Ritter, and Lu~Wang, editors, \emph{Proceedings of the 2025 Conference of the Nations of the Americas Chapter of the Association for Computational Linguistics: Human Language Technologies, {NAACL} 2025 - Volume 1: Long Papers, Albuquerque, New Mexico, USA, April 29 - May 4, 2025}, pages 10627--10646. Association for Computational Linguistics, 2025.
\newblock \doi{10.18653/V1/2025.NAACL-LONG.533}.
\newblock URL \url{https://doi.org/10.18653/v1/2025.naacl-long.533}.

\bibitem[Fu et~al.(2025)Fu, Gao, Shen, Zhu, Mei, He, Xu, Wei, Mei, Wang, Yang, Yuan, and Wu]{DBLP:journals/corr/abs-2505-24298}
Wei Fu, Jiaxuan Gao, Xujie Shen, Chen Zhu, Zhiyu Mei, Chuyi He, Shusheng Xu, Guo Wei, Jun Mei, Jiashu Wang, Tongkai Yang, Binhang Yuan, and Yi~Wu.
\newblock Areal: {A} large-scale asynchronous reinforcement learning system for language reasoning.
\newblock \emph{CoRR}, abs/2505.24298, 2025.
\newblock \doi{10.48550/ARXIV.2505.24298}.
\newblock URL \url{https://doi.org/10.48550/arXiv.2505.24298}.

\bibitem[Google(2024)]{GoogleGemini2024Update}
Google.
\newblock Introducing gemini 2.0: our new ai model for the agentic era, December 2024.
\newblock URL \url{https://blog.google/technology/google-deepmind/google-gemini-ai-update-december-2024/}.

\bibitem[G{\"{u}}l{\c{c}}ehre et~al.(2023)G{\"{u}}l{\c{c}}ehre, Paine, Srinivasan, Konyushkova, Weerts, Sharma, Siddhant, Ahern, Wang, Gu, Macherey, Doucet, Firat, and de~Freitas]{DBLP:journals/corr/abs-2308-08998}
{\c{C}}aglar G{\"{u}}l{\c{c}}ehre, Tom~Le Paine, Srivatsan Srinivasan, Ksenia Konyushkova, Lotte Weerts, Abhishek Sharma, Aditya Siddhant, Alex Ahern, Miaosen Wang, Chenjie Gu, Wolfgang Macherey, Arnaud Doucet, Orhan Firat, and Nando de~Freitas.
\newblock Reinforced self-training (rest) for language modeling.
\newblock \emph{CoRR}, abs/2308.08998, 2023.
\newblock \doi{10.48550/ARXIV.2308.08998}.
\newblock URL \url{https://doi.org/10.48550/arXiv.2308.08998}.

\bibitem[Guo et~al.(2025)Guo, Yang, Zhang, Song, Wang, Zhu, Xu, Zhang, Ma, Bi, Zhang, Yu, Wu, Wu, Gou, Shao, Li, Gao, Liu, Xue, Wang, Wu, Feng, Lu, Zhao, Deng, Ruan, Dai, Chen, Ji, Li, Lin, Dai, Luo, Hao, Chen, Li, Zhang, Xu, Ding, Gao, Qu, Li, Guo, Li, Chen, Yuan, Tu, Qiu, Li, Cai, Ni, Liang, Chen, Dong, Hu, You, Gao, Guan, Huang, Yu, Wang, Zhang, Zhao, Wang, Zhang, Xu, Xia, Zhang, Zhang, Tang, Zhou, Li, Wang, Li, Tian, Huang, Zhang, Wang, Chen, Du, Ge, Zhang, Pan, Wang, Chen, Jin, Chen, Lu, Zhou, Chen, Ye, Wang, Yu, Zhou, Pan, Li, Zhou, Wu, Yun, Pei, Sun, Wang, Zeng, Liu, Liang, Gao, Yu, Zhang, Xiao, An, Liu, Wang, Chen, Nie, Cheng, Liu, Xie, Liu, Yang, Li, Su, Lin, Li, Jin, Shen, Chen, Sun, Wang, Song, Zhou, Wang, Shan, Li, Wang, Wei, Zhang, Xu, Li, Zhao, Sun, Wang, Yu, Zhang, Shi, Xiong, He, Piao, Wang, Tan, Ma, Liu, Guo, Ou, Wang, Gong, Zou, He, Xiong, Luo, You, Liu, Zhou, Zhu, Huang, Li, Zheng, Zhu, Ma, Tang, Zha, Yan, Ren, Ren, Sha, Fu, Xu, Xie, Zhang, Hao, Ma, Yan, Wu, Gu, Zhu, Liu, Li, Xie, Song,
  Pan, Huang, Xu, Zhang, and Zhang]{DeepSeek-R1}
Daya Guo, Dejian Yang, Haowei Zhang, Junxiao Song, Peiyi Wang, Qihao Zhu, Runxin Xu, Ruoyu Zhang, Shirong Ma, Xiao Bi, Xiaokang Zhang, Xingkai Yu, Yu~Wu, Z.~F. Wu, Zhibin Gou, Zhihong Shao, Zhuoshu Li, Ziyi Gao, Aixin Liu, Bing Xue, Bingxuan Wang, Bochao Wu, Bei Feng, Chengda Lu, Chenggang Zhao, Chengqi Deng, Chong Ruan, Damai Dai, Deli Chen, Dongjie Ji, Erhang Li, Fangyun Lin, Fucong Dai, Fuli Luo, Guangbo Hao, Guanting Chen, Guowei Li, H.~Zhang, Hanwei Xu, Honghui Ding, Huazuo Gao, Hui Qu, Hui Li, Jianzhong Guo, Jiashi Li, Jingchang Chen, Jingyang Yuan, Jinhao Tu, Junjie Qiu, Junlong Li, J.~L. Cai, Jiaqi Ni, Jian Liang, Jin Chen, Kai Dong, Kai Hu, Kaichao You, Kaige Gao, Kang Guan, Kexin Huang, Kuai Yu, Lean Wang, Lecong Zhang, Liang Zhao, Litong Wang, Liyue Zhang, Lei Xu, Leyi Xia, Mingchuan Zhang, Minghua Zhang, Minghui Tang, Mingxu Zhou, Meng Li, Miaojun Wang, Mingming Li, Ning Tian, Panpan Huang, Peng Zhang, Qiancheng Wang, Qinyu Chen, Qiushi Du, Ruiqi Ge, Ruisong Zhang, Ruizhe Pan, Runji Wang, R.~J.
  Chen, R.~L. Jin, Ruyi Chen, Shanghao Lu, Shangyan Zhou, Shanhuang Chen, Shengfeng Ye, Shiyu Wang, Shuiping Yu, Shunfeng Zhou, Shuting Pan, S.~S. Li, Shuang Zhou, Shaoqing Wu, Tao Yun, Tian Pei, Tianyu Sun, T.~Wang, Wangding Zeng, Wen Liu, Wenfeng Liang, Wenjun Gao, Wenqin Yu, Wentao Zhang, W.~L. Xiao, Wei An, Xiaodong Liu, Xiaohan Wang, Xiaokang Chen, Xiaotao Nie, Xin Cheng, Xin Liu, Xin Xie, Xingchao Liu, Xinyu Yang, Xinyuan Li, Xuecheng Su, Xuheng Lin, X.~Q. Li, Xiangyue Jin, Xiaojin Shen, Xiaosha Chen, Xiaowen Sun, Xiaoxiang Wang, Xinnan Song, Xinyi Zhou, Xianzu Wang, Xinxia Shan, Y.~K. Li, Y.~Q. Wang, Y.~X. Wei, Yang Zhang, Yanhong Xu, Yao Li, Yao Zhao, Yaofeng Sun, Yaohui Wang, Yi~Yu, Yichao Zhang, Yifan Shi, Yiliang Xiong, Ying He, Yishi Piao, Yisong Wang, Yixuan Tan, Yiyang Ma, Yiyuan Liu, Yongqiang Guo, Yuan Ou, Yuduan Wang, Yue Gong, Yuheng Zou, Yujia He, Yunfan Xiong, Yuxiang Luo, Yuxiang You, Yuxuan Liu, Yuyang Zhou, Y.~X. Zhu, Yanping Huang, Yaohui Li, Yi~Zheng, Yuchen Zhu, Yunxian Ma, Ying
  Tang, Yukun Zha, Yuting Yan, Z.~Z. Ren, Zehui Ren, Zhangli Sha, Zhe Fu, Zhean Xu, Zhenda Xie, Zhengyan Zhang, Zhewen Hao, Zhicheng Ma, Zhigang Yan, Zhiyu Wu, Zihui Gu, Zijia Zhu, Zijun Liu, Zilin Li, Ziwei Xie, Ziyang Song, Zizheng Pan, Zhen Huang, Zhipeng Xu, Zhongyu Zhang, and Zhen Zhang.
\newblock Deepseek-r1 incentivizes reasoning in llms through reinforcement learning.
\newblock \emph{Nature}, 645\penalty0 (8081):\penalty0 633--638, 2025.
\newblock \doi{10.1038/s41586-025-09422-z}.
\newblock URL \url{https://doi.org/10.1038/s41586-025-09422-z}.

\bibitem[He et~al.(2025)He, Liu, Liu, Yan, Wang, Cheng, Zhang, Zhang, Xu, Shen, Li, Zeng, Wei, Cheng, An, Liu, and Zhou]{skywork_or1}
Jujie He, Jiacai Liu, Chris~Yuhao Liu, Rui Yan, Chaojie Wang, Peng Cheng, Xiaoyu Zhang, Fuxiang Zhang, Jiacheng Xu, Wei Shen, Siyuan Li, Liang Zeng, Tianwen Wei, Cheng Cheng, Bo~An, Yang Liu, and Yahui Zhou.
\newblock Skywork open reasoner 1 technical report.
\newblock \emph{CoRR}, abs/2505.22312, 2025.
\newblock \doi{10.48550/ARXIV.2505.22312}.
\newblock URL \url{https://doi.org/10.48550/arXiv.2505.22312}.

\bibitem[Jaech et~al.(2024)Jaech, Kalai, Lerer, Richardson, El{-}Kishky, Low, Helyar, Madry, Beutel, Carney, Iftimie, Karpenko, Passos, Neitz, Prokofiev, Wei, Tam, Bennett, Kumar, Saraiva, Vallone, Duberstein, Kondrich, Mishchenko, Applebaum, Jiang, Nair, Zoph, Ghorbani, Rossen, Sokolowsky, Barak, McGrew, Minaiev, Hao, Baker, Houghton, McKinzie, Eastman, Lugaresi, Bassin, Hudson, Li, de~Bourcy, Voss, Shen, Zhang, Koch, Orsinger, Hesse, Fischer, Chan, Roberts, Kappler, Levy, Selsam, Dohan, Farhi, Mely, Robinson, Tsipras, Li, Oprica, Freeman, Zhang, Wong, Proehl, Cheung, Mitchell, Wallace, Ritter, Mays, Wang, Such, Raso, Leoni, Tsimpourlas, Song, von Lohmann, Sulit, Salmon, Parascandolo, Chabot, Zhao, Brockman, Leclerc, Salman, Bao, Sheng, Andrin, Bagherinezhad, Ren, Lightman, Chung, Kivlichan, O'Connell, Osband, Gilaberte, and Akkaya]{openai_o1}
Aaron Jaech, Adam Kalai, Adam Lerer, Adam Richardson, Ahmed El{-}Kishky, Aiden Low, Alec Helyar, Aleksander Madry, Alex Beutel, Alex Carney, Alex Iftimie, Alex Karpenko, Alex~Tachard Passos, Alexander Neitz, Alexander Prokofiev, Alexander Wei, Allison Tam, Ally Bennett, Ananya Kumar, Andre Saraiva, Andrea Vallone, Andrew Duberstein, Andrew Kondrich, Andrey Mishchenko, Andy Applebaum, Angela Jiang, Ashvin Nair, Barret Zoph, Behrooz Ghorbani, Ben Rossen, Benjamin Sokolowsky, Boaz Barak, Bob McGrew, Borys Minaiev, Botao Hao, Bowen Baker, Brandon Houghton, Brandon McKinzie, Brydon Eastman, Camillo Lugaresi, Cary Bassin, Cary Hudson, Chak~Ming Li, Charles de~Bourcy, Chelsea Voss, Chen Shen, Chong Zhang, Chris Koch, Chris Orsinger, Christopher Hesse, Claudia Fischer, Clive Chan, Dan Roberts, Daniel Kappler, Daniel Levy, Daniel Selsam, David Dohan, David Farhi, David Mely, David Robinson, Dimitris Tsipras, Doug Li, Dragos Oprica, Eben Freeman, Eddie Zhang, Edmund Wong, Elizabeth Proehl, Enoch Cheung, Eric Mitchell,
  Eric Wallace, Erik Ritter, Evan Mays, Fan Wang, Felipe~Petroski Such, Filippo Raso, Florencia Leoni, Foivos Tsimpourlas, Francis Song, Fred von Lohmann, Freddie Sulit, Geoff Salmon, Giambattista Parascandolo, Gildas Chabot, Grace Zhao, Greg Brockman, Guillaume Leclerc, Hadi Salman, Haiming Bao, Hao Sheng, Hart Andrin, Hessam Bagherinezhad, Hongyu Ren, Hunter Lightman, Hyung~Won Chung, Ian Kivlichan, Ian O'Connell, Ian Osband, Ignasi~Clavera Gilaberte, and Ilge Akkaya.
\newblock Openai o1 system card.
\newblock \emph{CoRR}, abs/2412.16720, 2024.
\newblock \doi{10.48550/ARXIV.2412.16720}.
\newblock URL \url{https://doi.org/10.48550/arXiv.2412.16720}.

\bibitem[Liu et~al.(2025)Liu, Chen, Li, Qi, Pang, Du, Lee, and Lin]{DBLP:journals/corr/abs-2503-20783}
Zichen Liu, Changyu Chen, Wenjun Li, Penghui Qi, Tianyu Pang, Chao Du, Wee~Sun Lee, and Min Lin.
\newblock Understanding r1-zero-like training: {A} critical perspective.
\newblock \emph{CoRR}, abs/2503.20783, 2025.
\newblock \doi{10.48550/ARXIV.2503.20783}.
\newblock URL \url{https://doi.org/10.48550/arXiv.2503.20783}.

\bibitem[OpenAI(2025)]{openai_o3}
OpenAI.
\newblock Openai o3-mini system card, 2025.
\newblock URL \url{https://cdn.openai.com/o3-mini-system-card-feb10.pdf}.

\bibitem[Qwen(2025)]{qwq}
Qwen.
\newblock Qwq-32b: Embracing the power of reinforcement learning, March 2025.
\newblock URL \url{https://qwenlm.github.io/blog/qwq-32b/}.

\bibitem[Roux et~al.(2025)Roux, Bellemare, Lebensold, Bergeron, Greaves, Fr{\'{e}}chette, Pelletier, Thibodeau{-}Laufer, T{\'{o}}th, and Work]{DBLP:journals/corr/abs-2503-14286}
Nicolas~Le Roux, Marc~G. Bellemare, Jonathan Lebensold, Arnaud Bergeron, Joshua Greaves, Alexandre Fr{\'{e}}chette, Carolyne Pelletier, Eric Thibodeau{-}Laufer, S{\'{a}}ndor T{\'{o}}th, and Sam Work.
\newblock Tapered off-policy {REINFORCE:} stable and efficient reinforcement learning for llms.
\newblock \emph{CoRR}, abs/2503.14286, 2025.
\newblock \doi{10.48550/ARXIV.2503.14286}.
\newblock URL \url{https://doi.org/10.48550/arXiv.2503.14286}.

\bibitem[Schulman et~al.(2017)Schulman, Wolski, Dhariwal, Radford, and Klimov]{DBLP:journals/corr/SchulmanWDRK17}
John Schulman, Filip Wolski, Prafulla Dhariwal, Alec Radford, and Oleg Klimov.
\newblock Proximal policy optimization algorithms.
\newblock \emph{CoRR}, abs/1707.06347, 2017.
\newblock URL \url{http://arxiv.org/abs/1707.06347}.

\bibitem[Shao et~al.(2024)Shao, Wang, Zhu, Xu, Song, Zhang, Li, Wu, and Guo]{DBLP:journals/corr/abs-2402-03300e}
Zhihong Shao, Peiyi Wang, Qihao Zhu, Runxin Xu, Junxiao Song, Mingchuan Zhang, Y.~K. Li, Y.~Wu, and Daya Guo.
\newblock Deepseekmath: Pushing the limits of mathematical reasoning in open language models.
\newblock \emph{CoRR}, abs/2402.03300, 2024.
\newblock \doi{10.48550/ARXIV.2402.03300}.
\newblock URL \url{https://doi.org/10.48550/arXiv.2402.03300}.

\bibitem[Tang et~al.(2024)Tang, Guo, Zheng, Calandriello, Cao, Tarassov, Munos, Pires, Valko, Cheng, and Dabney]{DBLP:journals/corr/abs-2405-08448}
Yunhao Tang, Zhaohan~Daniel Guo, Zeyu Zheng, Daniele Calandriello, Yuan Cao, Eugene Tarassov, R{\'{e}}mi Munos, Bernardo~{\'{A}}vila Pires, Michal Valko, Yong Cheng, and Will Dabney.
\newblock Understanding the performance gap between online and offline alignment algorithms.
\newblock \emph{CoRR}, abs/2405.08448, 2024.
\newblock \doi{10.48550/ARXIV.2405.08448}.
\newblock URL \url{https://doi.org/10.48550/arXiv.2405.08448}.

\bibitem[Team et~al.(2025)Team, Du, Gao, Xing, Jiang, Chen, Li, Xiao, Du, Liao, Tang, Wang, Zhang, Yuan, Lu, Tang, Sung, Wei, Lai, Guo, Zhu, Ding, Hu, Yang, Zhang, Yao, Zhao, Lu, Li, Yu, Gao, Zheng, Yuan, Chen, Guo, Su, Wang, Zhao, Zhang, Liu, Yan, Wu, Shi, Ye, Yu, Dong, Zhang, Ma, Pan, Gong, Liu, Ma, Wei, Cao, Huang, Jiang, Gao, Xiong, He, Huang, Wu, He, Wei, Jia, Wu, Xu, Zu, Zhou, Pan, Charles, Li, Hu, Liu, Chen, Wang, Liu, Qin, Liu, Yang, Bao, Du, Wu, Wang, Zhou, Wang, Li, Zhu, Zhang, Wang, Yang, Huang, Huang, Xu, and Yang]{DBLP:journals/corr/abs-2501-12599}
Kimi Team, Angang Du, Bofei Gao, Bowei Xing, Changjiu Jiang, Cheng Chen, Cheng Li, Chenjun Xiao, Chenzhuang Du, Chonghua Liao, Chuning Tang, Congcong Wang, Dehao Zhang, Enming Yuan, Enzhe Lu, Fengxiang Tang, Flood Sung, Guangda Wei, Guokun Lai, Haiqing Guo, Han Zhu, Hao Ding, Hao Hu, Hao Yang, Hao Zhang, Haotian Yao, Haotian Zhao, Haoyu Lu, Haoze Li, Haozhen Yu, Hongcheng Gao, Huabin Zheng, Huan Yuan, Jia Chen, Jianhang Guo, Jianlin Su, Jianzhou Wang, Jie Zhao, Jin Zhang, Jingyuan Liu, Junjie Yan, Junyan Wu, Lidong Shi, Ling Ye, Longhui Yu, Mengnan Dong, Neo Zhang, Ningchen Ma, Qiwei Pan, Qucheng Gong, Shaowei Liu, Shengling Ma, Shupeng Wei, Sihan Cao, Siying Huang, Tao Jiang, Weihao Gao, Weimin Xiong, Weiran He, Weixiao Huang, Wenhao Wu, Wenyang He, Xianghui Wei, Xianqing Jia, Xingzhe Wu, Xinran Xu, Xinxing Zu, Xinyu Zhou, Xuehai Pan, Y.~Charles, Yang Li, Yangyang Hu, Yangyang Liu, Yanru Chen, Yejie Wang, Yibo Liu, Yidao Qin, Yifeng Liu, Ying Yang, Yiping Bao, Yulun Du, Yuxin Wu, Yuzhi Wang, Zaida Zhou,
  Zhaoji Wang, Zhaowei Li, Zhen Zhu, Zheng Zhang, Zhexu Wang, Zhilin Yang, Zhiqi Huang, Zihao Huang, Ziyao Xu, and Zonghan Yang.
\newblock Kimi k1.5: Scaling reinforcement learning with llms.
\newblock \emph{CoRR}, abs/2501.12599, 2025.
\newblock \doi{10.48550/ARXIV.2501.12599}.
\newblock URL \url{https://doi.org/10.48550/arXiv.2501.12599}.

\bibitem[Trung et~al.(2024)Trung, Zhang, Jie, Sun, Jin, and Li]{DBLP:conf/acl/TrungZJSJL24}
Luong~Quoc Trung, Xinbo Zhang, Zhanming Jie, Peng Sun, Xiaoran Jin, and Hang Li.
\newblock Reft: Reasoning with reinforced fine-tuning.
\newblock In Lun{-}Wei Ku, Andre Martins, and Vivek Srikumar, editors, \emph{Proceedings of the 62nd Annual Meeting of the Association for Computational Linguistics (Volume 1: Long Papers), {ACL} 2024, Bangkok, Thailand, August 11-16, 2024}, pages 7601--7614. Association for Computational Linguistics, 2024.
\newblock \doi{10.18653/V1/2024.ACL-LONG.410}.
\newblock URL \url{https://doi.org/10.18653/v1/2024.acl-long.410}.

\bibitem[Wang et~al.(2025{\natexlab{a}})Wang, Yu, Gao, Zheng, Liu, Lu, Dang, Chen, Yang, Zhang, Liu, Yang, Zhao, Yue, Song, Yu, Huang, and Lin]{DBLP:journals/corr/abs-2506-01939}
Shenzhi Wang, Le~Yu, Chang Gao, Chujie Zheng, Shixuan Liu, Rui Lu, Kai Dang, Xionghui Chen, Jianxin Yang, Zhenru Zhang, Yuqiong Liu, An~Yang, Andrew Zhao, Yang Yue, Shiji Song, Bowen Yu, Gao Huang, and Junyang Lin.
\newblock Beyond the 80/20 rule: High-entropy minority tokens drive effective reinforcement learning for {LLM} reasoning.
\newblock \emph{CoRR}, abs/2506.01939, 2025{\natexlab{a}}.
\newblock \doi{10.48550/ARXIV.2506.01939}.
\newblock URL \url{https://doi.org/10.48550/arXiv.2506.01939}.

\bibitem[Wang et~al.(2025{\natexlab{b}})Wang, Zhou, Li, and Liu]{DBLP:journals/corr/abs-2506-20512}
Zengzhi Wang, Fan Zhou, Xuefeng Li, and Pengfei Liu.
\newblock Octothinker: Mid-training incentivizes reinforcement learning scaling.
\newblock \emph{CoRR}, abs/2506.20512, 2025{\natexlab{b}}.
\newblock \doi{10.48550/ARXIV.2506.20512}.
\newblock URL \url{https://doi.org/10.48550/arXiv.2506.20512}.

\bibitem[Wen et~al.(2025)Wen, Cai, Xiao, He, An, Duan, Du, Liu, Tang, Lv, Zou, Deng, Jia, and Zhang]{DBLP:journals/corr/abs-2503-10460}
Liang Wen, Yunke Cai, Fenrui Xiao, Xin He, Qi~An, Zhenyu Duan, Yimin Du, Junchen Liu, Lifu Tang, Xiaowei Lv, Haosheng Zou, Yongchao Deng, Shousheng Jia, and Xiangzheng Zhang.
\newblock Light-r1: Curriculum sft, {DPO} and {RL} for long {COT} from scratch and beyond.
\newblock \emph{CoRR}, abs/2503.10460, 2025.
\newblock \doi{10.48550/ARXIV.2503.10460}.
\newblock URL \url{https://doi.org/10.48550/arXiv.2503.10460}.

\bibitem[Williams(1992)]{DBLP:journals/ml/Williams92}
Ronald~J. Williams.
\newblock Simple statistical gradient-following algorithms for connectionist reinforcement learning.
\newblock \emph{Mach. Learn.}, 8:\penalty0 229--256, 1992.
\newblock \doi{10.1007/BF00992696}.
\newblock URL \url{https://doi.org/10.1007/BF00992696}.

\bibitem[Xi et~al.(2024)Xi, Chen, Hong, Jin, Zheng, He, Ding, Liu, Guo, Wang, Guo, Shen, Fan, Zhou, Dou, Wang, Zhang, Sun, Gui, Zhang, and Huang]{DBLP:conf/icml/XiCHJZHDLGWGSFZ24}
Zhiheng Xi, Wenxiang Chen, Boyang Hong, Senjie Jin, Rui Zheng, Wei He, Yiwen Ding, Shichun Liu, Xin Guo, Junzhe Wang, Honglin Guo, Wei Shen, Xiaoran Fan, Yuhao Zhou, Shihan Dou, Xiao Wang, Xinbo Zhang, Peng Sun, Tao Gui, Qi~Zhang, and Xuanjing Huang.
\newblock Training large language models for reasoning through reverse curriculum reinforcement learning.
\newblock In \emph{Forty-first International Conference on Machine Learning, {ICML} 2024, Vienna, Austria, July 21-27, 2024}. OpenReview.net, 2024.
\newblock URL \url{https://openreview.net/forum?id=t82Y3fmRtk}.

\bibitem[Yang et~al.(2024)Yang, Zhang, Hui, Gao, Yu, Li, Liu, Tu, Zhou, Lin, Lu, Xue, Lin, Liu, Ren, and Zhang]{DBLP:journals/corr/abs-2409-12122}
An~Yang, Beichen Zhang, Binyuan Hui, Bofei Gao, Bowen Yu, Chengpeng Li, Dayiheng Liu, Jianhong Tu, Jingren Zhou, Junyang Lin, Keming Lu, Mingfeng Xue, Runji Lin, Tianyu Liu, Xingzhang Ren, and Zhenru Zhang.
\newblock Qwen2.5-math technical report: Toward mathematical expert model via self-improvement.
\newblock \emph{CoRR}, abs/2409.12122, 2024.
\newblock \doi{10.48550/ARXIV.2409.12122}.
\newblock URL \url{https://doi.org/10.48550/arXiv.2409.12122}.

\bibitem[Yang et~al.(2025{\natexlab{a}})Yang, Li, Yang, Zhang, Hui, Zheng, Yu, Gao, Huang, Lv, Zheng, Liu, Zhou, Huang, Hu, Ge, Wei, Lin, Tang, Yang, Tu, Zhang, Yang, Yang, Zhou, Lin, Dang, Bao, Yang, Yu, Deng, Li, Xue, Li, Zhang, Wang, Zhu, Men, Gao, Liu, Luo, Li, Tang, Yin, Ren, Wang, Zhang, Ren, Fan, Su, Zhang, Zhang, Wan, Liu, Wang, Cui, Zhang, Zhou, and Qiu]{DBLP:journals/corr/abs-2505-09388}
An~Yang, Anfeng Li, Baosong Yang, Beichen Zhang, Binyuan Hui, Bo~Zheng, Bowen Yu, Chang Gao, Chengen Huang, Chenxu Lv, Chujie Zheng, Dayiheng Liu, Fan Zhou, Fei Huang, Feng Hu, Hao Ge, Haoran Wei, Huan Lin, Jialong Tang, Jian Yang, Jianhong Tu, Jianwei Zhang, Jian Yang, Jiaxi Yang, Jingren Zhou, Junyang Lin, Kai Dang, Keqin Bao, Kexin Yang, Le~Yu, Lianghao Deng, Mei Li, Mingfeng Xue, Mingze Li, Pei Zhang, Peng Wang, Qin Zhu, Rui Men, Ruize Gao, Shixuan Liu, Shuang Luo, Tianhao Li, Tianyi Tang, Wenbiao Yin, Xingzhang Ren, Xinyu Wang, Xinyu Zhang, Xuancheng Ren, Yang Fan, Yang Su, Yichang Zhang, Yinger Zhang, Yu~Wan, Yuqiong Liu, Zekun Wang, Zeyu Cui, Zhenru Zhang, Zhipeng Zhou, and Zihan Qiu.
\newblock Qwen3 technical report.
\newblock \emph{CoRR}, abs/2505.09388, 2025{\natexlab{a}}.
\newblock \doi{10.48550/ARXIV.2505.09388}.
\newblock URL \url{https://doi.org/10.48550/arXiv.2505.09388}.

\bibitem[Yang et~al.(2025{\natexlab{b}})Yang, Dou, Guo, Lu, Ju, Deng, and Xin]{yang2025dcpo}
Shihui Yang, Chengfeng Dou, Peidong Guo, Kai Lu, Qiang Ju, Fei Deng, and Rihui Xin.
\newblock Dcpo: Dynamic clipping policy optimization.
\newblock \emph{arXiv preprint arXiv:2509.02333}, 2025{\natexlab{b}}.

\bibitem[Yang et~al.(2025{\natexlab{c}})Yang, Luo, Wang, Han, He, Li, and Xu]{DBLP:journals/corr/abs-2505-12929}
Zhihe Yang, Xufang Luo, Zilong Wang, Dongqi Han, Zhiyuan He, Dongsheng Li, and Yunjian Xu.
\newblock Do not let low-probability tokens over-dominate in {RL} for llms.
\newblock \emph{CoRR}, abs/2505.12929, 2025{\natexlab{c}}.
\newblock \doi{10.48550/ARXIV.2505.12929}.
\newblock URL \url{https://doi.org/10.48550/arXiv.2505.12929}.

\bibitem[Yu et~al.(2025)Yu, Zhang, Zhu, Yuan, Zuo, Yue, Fan, Liu, Liu, Liu, Lin, Lin, Ma, Sheng, Tong, Zhang, Zhang, Zhang, Zhu, Zhu, Chen, Chen, Wang, Yu, Dai, Song, Wei, Zhou, Liu, Ma, Zhang, Yan, Qiao, Wu, and Wang]{DBLP:journals/corr/abs-2503-14476}
Qiying Yu, Zheng Zhang, Ruofei Zhu, Yufeng Yuan, Xiaochen Zuo, Yu~Yue, Tiantian Fan, Gaohong Liu, Lingjun Liu, Xin Liu, Haibin Lin, Zhiqi Lin, Bole Ma, Guangming Sheng, Yuxuan Tong, Chi Zhang, Mofan Zhang, Wang Zhang, Hang Zhu, Jinhua Zhu, Jiaze Chen, Jiangjie Chen, Chengyi Wang, Hongli Yu, Weinan Dai, Yuxuan Song, Xiangpeng Wei, Hao Zhou, Jingjing Liu, Wei{-}Ying Ma, Ya{-}Qin Zhang, Lin Yan, Mu~Qiao, Yonghui Wu, and Mingxuan Wang.
\newblock {DAPO:} an open-source {LLM} reinforcement learning system at scale.
\newblock \emph{CoRR}, abs/2503.14476, 2025.
\newblock \doi{10.48550/ARXIV.2503.14476}.
\newblock URL \url{https://doi.org/10.48550/arXiv.2503.14476}.

\bibitem[Yuan et~al.(2025)Yuan, Xiao, Leng, Wang, Li, Xu, Chan, Zhao, Xu, Wei, Zhang, and Rong]{DBLP:journals/corr/abs-2507-22607}
Ruifeng Yuan, Chenghao Xiao, Sicong Leng, Jianyu Wang, Long Li, Weiwen Xu, Hou~Pong Chan, Deli Zhao, Tingyang Xu, Zhongyu Wei, Hao Zhang, and Yu~Rong.
\newblock Vl-cogito: Progressive curriculum reinforcement learning for advanced multimodal reasoning.
\newblock \emph{CoRR}, abs/2507.22607, 2025.
\newblock \doi{10.48550/ARXIV.2507.22607}.
\newblock URL \url{https://doi.org/10.48550/arXiv.2507.22607}.

\bibitem[Zhang et~al.(2025)Zhang, Zuo, He, Sun, Liu, Jiang, Fan, Tian, Jia, Li, et~al.]{zhang2025survey}
Kaiyan Zhang, Yuxin Zuo, Bingxiang He, Youbang Sun, Runze Liu, Che Jiang, Yuchen Fan, Kai Tian, Guoli Jia, Pengfei Li, et~al.
\newblock A survey of reinforcement learning for large reasoning models.
\newblock \emph{arXiv preprint arXiv:2509.08827}, 2025.

\bibitem[Zheng et~al.(2025)Zheng, Liu, Li, Chen, Yu, Gao, Dang, Liu, Men, Yang, Zhou, and Lin]{DBLP:journals/corr/abs-2507-18071}
Chujie Zheng, Shixuan Liu, Mingze Li, Xiong{-}Hui Chen, Bowen Yu, Chang Gao, Kai Dang, Yuqiong Liu, Rui Men, An~Yang, Jingren Zhou, and Junyang Lin.
\newblock Group sequence policy optimization.
\newblock \emph{CoRR}, abs/2507.18071, 2025.
\newblock \doi{10.48550/ARXIV.2507.18071}.
\newblock URL \url{https://doi.org/10.48550/arXiv.2507.18071}.

\bibitem[Zhu et~al.(2025)Zhu, Cheng, Zhang, Li, Zhang, Jiang, Sun, Hua, Zuo, Lv, et~al.]{zhu2025flowrl}
Xuekai Zhu, Daixuan Cheng, Dinghuai Zhang, Hengli Li, Kaiyan Zhang, Che Jiang, Youbang Sun, Ermo Hua, Yuxin Zuo, Xingtai Lv, et~al.
\newblock Flowrl: Matching reward distributions for llm reasoning.
\newblock \emph{arXiv preprint arXiv:2509.15207}, 2025.

\end{thebibliography}

\clearpage
\newpage

\appendix
\section*{\centering \LARGE{Appendix}}

\section{Performance on OctoThinker-Llama}\label{appendix:llama}
\begin{figure}[ht]
\begin{center}
\includegraphics[width=0.9\linewidth]{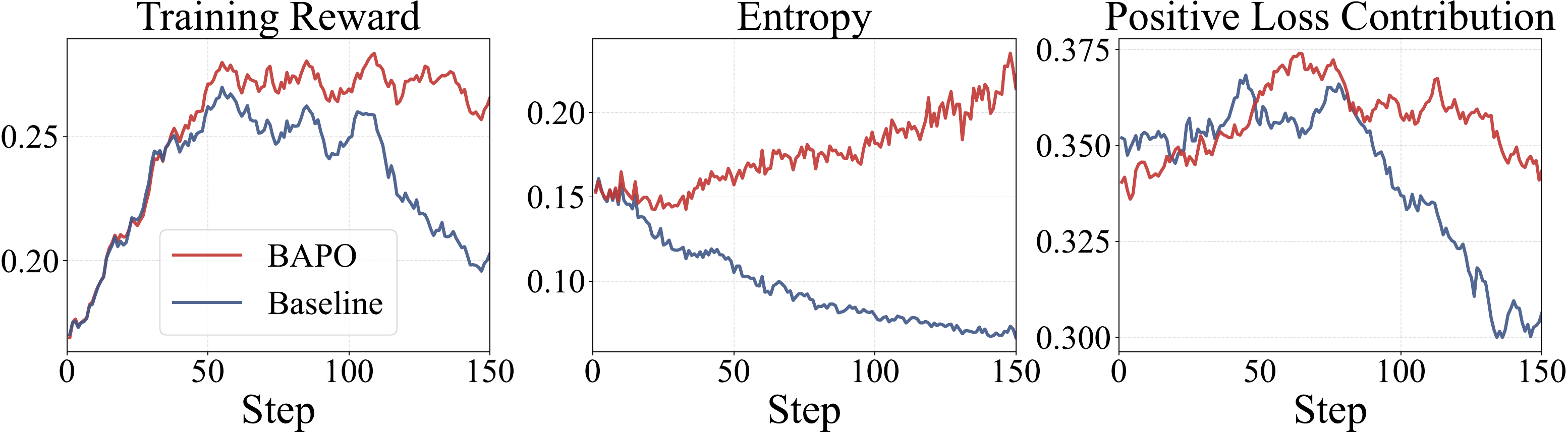}
\end{center}
\caption{Training dynamics of OctoThinker-Llama-3B-Long-Zero.}
\label{fig:Llama_training_dynamics}
\end{figure}

We illustrate the training dynamics on  OctoThinker-Llama in Figure \ref{fig:Llama_training_dynamics}. Since Llama family models behave badly in RL training, we choose the model after mid-training \citep{DBLP:journals/corr/abs-2506-20512} to show the robustness of BAPO. We can find that BAPO provides consistent and significant improvement in training. For training details, we set the low bound as $0.8$-$0.9$, high bound as $1.2$-$2.0$, and target positive loss contribution as $0.45$.

\section{Proofs of Equation \ref{equation:entropy} }\label{appendix:proof}

\subsection{Explanations for all variables and expressions}

All notation used in the following justification, including variables and expressions, is provided with detailed explanations in Table \ref{tab:definition}.

\begin{table*}[!ht]
\centering
\caption{Notation used in justification below.}
\resizebox{0.99\textwidth}{!}{ 
\renewcommand{\arraystretch}{2}
\begin{tabular}{lll}
\toprule
\textbf{Category} & \textbf{Symbol} & \textbf{Meaning} \\
\midrule
\multirow{6}{*}{Variables} 
  & $\pi_{\theta}$ & The policy parameterized by $\theta$ \\
  & $\pi_{\theta_{rollout}}$ & The standard sampling policy \\
  & $\bm{x}$ & Given prompt \\
  & $\bm{y}$ & A T-token response generated  by $\pi_{\theta}$ when given $\bm{x}$ \\
  & $y_t$ & The t-th token of y \\
  & $\eta$ & Learning rate \\
\midrule
\multirow{15}{*}{Expressions} 
& $\pi_{\theta}(\cdot|\bm{x},\bm{y}_{<t})$ & 
\parbox[t]{0.6\textwidth}{Probability of generating token $\cdot$ under policy $\pi_{\theta}$ given input $\bm{x}$ and previous tokens $\bm{y}_{<t}$} \\
& $\pi_{\theta_{rollout}}(y_t|\bm{x},\bm{y}_{<t})$ & 
\parbox[t]{0.6\textwidth}{Probability of generating token $\cdot$ under standard sampling policy $\pi_{\theta_{rollout}}$ given input $\bm{x}$ and previous tokens $\bm{y}_{<t}$} \\
& $A(\cdot|\bm{x},\bm{y_{<t}})$ & \parbox[t]{0.6\textwidth}{The measurement of how much better(or worse) selecting token $\cdot$ is compared to the expected value under the current policy, given $\bm{x}$ and $\bm{y_{<t}}$} \\
& $\mathcal{H}(\cdot|\bm{x},\bm{y}_{<t})$ & \parbox[t]{0.6\textwidth}{The information entropy of policy $\cdot$ given $\bm{x}$ and $\bm{y}_{<t}$}\\
& $Cov_{y_t \sim \pi_{\theta}(\cdot|\bm{x},\bm{y}_{<t})}(a(y_t),b(y_t))$ & \parbox[t]{0.6\textwidth}{The expected covariance of $a(y_t)$ and $b(y_t)$ over $y_t$ sampled from the policy $\pi_{\theta}$, given $\bm{x}$ and $\bm{y}_{<t}$} \\
& $\mathbb{I}(a=b)$ & \parbox[t]{0.6\textwidth}{Indicator function that equals 1 if $a=b$ and 0 otherwise} \\
& $Q^{(\pi_{\theta})}(\cdot,\bm{x})$ & \parbox[t]{0.6\textwidth}{The expected cumulative reward obtained by taking token $\cdot$ given input $\bm{x}$ and previous tokens under policy $\pi_{\theta}$} \\
& $V^{(\pi_{\theta})}(\bm{x})$ & \parbox[t]{0.6\textwidth}{The expected return of the new taking token given input $\bm{x}$ and previous tokens under policy $\pi_{\theta}$} \\
& $z_{\bm{y},\bm{x}}$ & \parbox[t]{0.6\textwidth}{A quantity representing the cumulative weight of sequence $\bm{y}$ given input $\bm{x}$ under policy $\pi_\theta$, reflecting its contribution to the policy taken at the current optimization step} \\

& $\nabla_{\theta_{y_t,\bm{x}}} J(\theta)$ & \parbox[t]{0.6\textwidth}{The gradient of the policy taken with respect to the logit parameter $\theta_{y_t,\bm{x}}$, representing how the policy $\pi_\theta$ should be adjusted for token $y_t$ given input $\bm{x}$} \\
\bottomrule
\end{tabular}
}
\label{tab:definition}
\end{table*}

\subsection{Preparation: Rewrite the PPO derivatives}
To facilitate the justification of the propositions below, we rewrite the PPO loss function in the following form:

\begin{align*}
\nabla J^{\textnormal{PPO}} &= 
\underbrace{\sum_{A(y_t)>0} \pi_{\theta}(y_t) \cdot \mathbb{I}\{r(y_t)<1+\varepsilon\} \cdot A(y_t) \cdot \nabla \log \pi_\theta(y_t)}_{\textnormal{positive tokens}} \\
&\quad + 
\underbrace{\sum_{A(y_t)<0} \pi_{\theta}(y_t) \cdot \mathbb{I}\{r(y_t)>1-\varepsilon\} \cdot A(y_t) \cdot \nabla \log \pi_\theta(y_t)}_{\textnormal{negative tokens}}
\end{align*}

where \[
\pi_{\theta}(y_t) = \pi_{\theta}(y_t|\bm{x},\bm{y_{<t}})\ ,\ r(y_t) = \frac{\pi_\theta(y_t \mid \bm{x}, \bm{y_{<t}})}{\pi_{\theta_\text{rollout}}(y_t \mid \bm{x},\bm{y_{<t}})}\ ,\ A(y_t) = A(y_t|\bm{x},\bm{y_{<t}}) \ .
\]

\subsection{Proofs of the main Propositions}

The following derivation is inspired by the proof framework in \citet{DBLP:journals/corr/abs-2505-22617}. While the original work focuses mainly on the  basic gradient formulation of naive REINFORCE to provide a heuristic explanation, our study advances this approach by deriving the gradient expression \textbf{specific to the PPO objective}. This refinement offers a specific, \textbf{intuitive  yet theoretical} account of how policy entropy is intrinsically shaped by the interaction between token-level advantages and their sampling probabilities.

\subsubsection{Preclaims}

\noindent {Proofs of these three lemmas below are available in \citet{DBLP:journals/corr/abs-2505-22617}.}

\begin{lemma}
Let the actor policy $\pi_\theta$ be a tabular softmax policy, 
the difference of information entropy given prompt $x$ between two consecutive steps $k$ and $k+1$ satisfies
\[
\mathcal{H}(\pi_\theta^{k+1}|\bm{x},\bm{y_{<t}}) - \mathcal{H}(\pi_\theta^k|\bm{x},\bm{y_{<t}}) 
\approx - \operatorname{Cov}_{y_t \sim \pi_\theta^k(\cdot|\bm{x},\bm{y_{<t}})}
\left( \log \pi_\theta^k(y_t), \; z_{ \bm{y},\bm{x}}^{k+1} - z_{\bm{y},\bm{x}}^k \right).
\]

\end{lemma}

\begin{lemma}[Derivative of softmax function]
\[
\frac{\partial \log \pi_\theta(y_t)}{\partial \theta_{y_t',\bm{x}}} 
= \mathbb{I}\{y_t=y_t'\} - \pi_\theta(y_t')
\]

\end{lemma}

\begin{lemma}[Expectation of Advantage function given prompt $x$]
\[ \begin{aligned} \mathbb{E}_{y_t \sim \pi_\theta(\cdot|x,\bm{y_{<t}})} \big[ A^{\pi_\theta}(y_t) \big] &= \mathbb{E}_{y_t \sim \pi_\theta(\cdot|x,\bm{y_{<t}})} \big[ Q^{\pi_\theta}(y_t,\bm{x}) - V^{\pi_\theta}(\bm{x}) \big] \\ &= \mathbb{E}_{y_t \sim \pi_\theta(\cdot|x,\bm{y_{<t}})} \big[ Q(y_t,\bm{x}) \big] - \mathbb{E}_{y_t \sim \pi_\theta(\cdot|x,\bm{y_{<t}})} \big[ V(\bm{x}) \big] \\ &= V(\bm{x}) - V(\bm{x}) \\ &= 0 \end{aligned} \]

\end{lemma}

\subsubsection{Principle Propositions}

\textbf{Proposition 1:} Assume the actor policy $\pi_\theta$ follows a tabular softmax policy and is optimized via the PPO objective, the difference of $z_{\bm{y},\bm{x}}$ between two consecutive steps k and k+1 satisfies
\begin{equation*}
z_{\bm{y},\bm{x}}^{k+1} - z_{\bm{y},\bm{x}}^k = \eta \cdot \pi_\theta(y_t) \cdot [A(y_t) \cdot \mathcal{X}(y_t)+C],
\end{equation*}

where 
\[
\mathcal{X}(y_t) =
\begin{cases}
1, & \bm{\mathit{if}}\; A(y_t) > 0 \ \&\  r(y_t)< 1+\epsilon \\
& \bm{\mathit{or}}\; A(y_t) < 0 \ \&\ r(y_t)>1-\epsilon \\
0, & \text{otherwise}
\end{cases} 
\]

and $C$ includes all clauses irrelevant to $y_t$.

\noindent\textbf{{It is worth noting that $\bm{\mathcal{X}(y_t) = 0}$ if and only if $\bm{y_t}$ is clipped.}}

\begin{proof}

In tabular softmax policy, each trajectory-prompt pair $(\bm{y},\bm{x})$ is associated with an individual logit parameter $z_{\bm{y},\bm{x}} = \theta_{y_t,\bm{x}}$. Through gradient backtracking, $z_{\bm{y},\bm{x}}$ is updated via $z_{\bm{y},\bm{x}}^{k+1} = z_{\bm{y},\bm{x}}^k + \eta \cdot \nabla_{\theta_{y_t,\bm{x}}} J(\theta)$. According to the loss function of PPO, we have

\begin{align*}
z_{\bm{y},\bm{x}}^{k+1} - z_{\bm{y},\bm{x}}^k &= \eta \cdot \nabla_{\theta_{y_t,\bm{x}}} J_{PPO}(\theta) \\
&= \eta \cdot \mathbb{E}_{\substack{y_t' \sim \pi_\theta(\cdot|\bm{x},\bm{y_{<t}}) \\ A(y_t') > 0}} \left[ \mathbb{I}\{r(y_t')<1+\varepsilon\} \cdot \nabla_{\theta_{y_t,\bm{x}}} \log \pi_\theta(y_t') \cdot A(y_t') \right] \\
&+ \eta \cdot \mathbb{E}_{\substack{y_t' \sim \pi_\theta(\cdot|\bm{x},\bm{y_{<t}}) \\ A(y_t') < 0}} \left[ \mathbb{I}\{r(y_t')>1-\varepsilon\} \cdot \nabla_{\theta_{y_t,\bm{x}}} \log \pi_\theta(y_t') \cdot A(y_t') \right] \\
&= \underbrace{\eta \cdot \mathbb{E}_{y_t' \sim \pi_\theta(\cdot|\bm{x},\bm{y_{<t}})} \left[ \nabla_{\theta_{y_t,\bm{x}}} \log \pi_\theta(y_t') \cdot A(y_t') \right]}_{\Circled{1}} \\
&- \underbrace{\eta \cdot \mathbb{E}_{\substack{y_t' \sim \pi_\theta(\cdot|\bm{x},\bm{y_{<t}}) \\ A(y_t') > 0}} \left[ \mathbb{I}\{r(y_t')>1+\varepsilon\} \cdot \nabla_{\theta_{y_t,\bm{x}}} \log \pi_\theta(y_t') \cdot A(y_t') \right]}_{\Circled{2}} \\
&- \underbrace{\eta \cdot \mathbb{E}_{\substack{y_t' \sim \pi_\theta(\cdot|\bm{x},\bm{y_{<t}}) \\ A(y_t') < 0}} \left[ \mathbb{I}\{r(y_t')<1-\varepsilon\} \cdot \nabla_{\theta_{y_t.\bm{x}}} \log \pi_\theta(y_t') \cdot A(y_t') \right]}_{\Circled{3}}\\
&= \Circled{1}-(\Circled{2}+\Circled{3}) \tag{8}
\end{align*}

We first perform the derivation on the term marked as \Circled{1}:

\begin{align*}
\Circled{1}&=\eta \cdot \mathbb{E}_{y_t' \sim \pi_\theta(\cdot|\bm{x},\bm{y_{<t}})} \left[ {\frac{\partial \log \pi_\theta(y_t')}{\partial \theta_{y_t,\bm{x}}}} \cdot A(y_t') \right] \\
&\overset{\text{Lemma 2}}{=}  \eta \cdot \sum_{y_t'} \left[ \pi_\theta(y_t') \cdot (\mathbb{I}\{y_t'=y_t\} - \pi_\theta(y_t)) \cdot A(y_t') \right] \\
&= \eta \cdot \pi_\theta(y_t) \cdot \left[ (1 - \pi_\theta(y_t)) \cdot A(y_t) - \sum_{ y_t' \neq y_t} \pi_\theta(y_t') \cdot A(y_t') \right] \\
&= \eta \cdot \pi_\theta(y_t) \cdot \left[ A(y_t) - \sum_{y_t'} \pi_\theta(y_t') \cdot A(y_t') \right] \\
&\overset{\text{Lemma 3}}{=}  \eta \cdot \pi_\theta(y_t) \cdot [A(y_t) - 0] \\
&= \eta \cdot \pi_\theta(y_t) \cdot A(y_t)
\end{align*}

To keep the presentation concise, we provide only the resulting derivations of Term \Circled{2} and \Circled{3}, as the detailed steps follow similarly to those for Term \Circled{1}.

\begin{align*}
\Circled{2}+\Circled{3}&=\eta \cdot \pi_\theta(y_t) \cdot A(y_t) \cdot (1-\mathcal{X}(y_t)) \\
&- \eta \cdot \pi_\theta(y_t) \cdot \sum_{A(y_t')>0} \left[ \mathbb{I}\{r(y_t')>1+\varepsilon\} \cdot  \pi_\theta(y_t') \cdot A(y_t') \right]\\
&-\eta \cdot \pi_\theta(y_t) \cdot \sum_{A(y_t')<0} \left[ \mathbb{I}\{r(y_t')<1-\varepsilon\} \cdot  \pi_\theta(y_t') \cdot A(y_t') \right]
\end{align*}

By substituting the results of the above derivation into Clause (8), we observe that:
\begin{align*}
(8)&= \Circled{1}-(\Circled{2}+\Circled{3}) \\
&=\eta \cdot \pi_\theta(y_t) \cdot \Bigl\{ A(y_t) \cdot \mathcal{X}(y_t) \\
&+ \sum_{A(y_t')>0} \left[ \mathbb{I}\{r(y_t')>1+\varepsilon\} \cdot  \pi_\theta(y_t') \cdot A(y_t') \right]\\
&+ \sum_{A(y_t')<0} \left[ \mathbb{I}\{r(y_t')<1-\varepsilon\} \cdot  \pi_\theta(y_t') \cdot A(y_t') \right]\Bigr\}
\end{align*}

By grouping all elements unrelated to $y_t$ into $C$, we are able to successfully establish our proposition.

\end{proof}

Building on Proposition 1, we establish the relationship between policy entropy and the covariance of specific tokens, which is stated as Proposition 2 below.\\

\noindent\textbf{Proposition 2 (Equation \ref{equation:entropy}):} Let the actor policy $\pi_\theta$ be tabular softmax policy, and $\pi_\theta$ is updated via PPO objective, the difference of information entropy given prompt $x$ and trajectory part $y_{<t}$ between two consecutive steps k and k+1 satisfies
\begin{equation*}
\mathcal{H}(\pi_\theta^{k+1}|\bm{x},\bm{y_{<t}}) - \mathcal{H}(\pi_\theta^k|\bm{x},\bm{y_{<t}})  \approx -\eta \cdot \text{Cov}_{y_t \sim \pi_\theta^k(\cdot|\bm{x},\bm{y_{<t}})} \left( \log \pi_\theta^k(y_t), A(y_t) \cdot \mathcal{X}(y_t)+C \right).
\end{equation*}\label{covariance equa}

\begin{proof}
Leveraging the conclusions of Lemma 1 and Proposition 1, we find that, under policy optimization and iteration via the PPO algorithm, the following relationship is satisfied:

\begin{equation*}
z_{\bm{y},\bm{x}}^{k+1} - z_{\bm{y},\bm{x}}^k = \eta \cdot (A(y_t) \cdot \mathcal{X}(y_t) +C).
\end{equation*}

Applying this into Lemma 1, we have

\begin{equation*}
\mathcal{H}(\pi_\theta^{k+1}|\bm{x},\bm{y_{<t}}) - \mathcal{H}(\pi_\theta^k|\bm{x},\bm{y_{<t}}) \approx -\eta \cdot \text{Cov}_{y_t \sim \pi_\theta^k(\cdot|\bm{x},\bm{y_{<t}})} \left( \log \pi_\theta^k(y_t), A(y_t) \cdot \mathcal{X}(y_t) +C \right).
\end{equation*}

\end{proof}

\subsection{Analysis}

\subsubsection{Direct Analysis: Why Varying \texorpdfstring{$\varepsilon$}{epsilon} Alters Entropy?}
We begin by examining the covariance of the clipped token, denoted as $\alpha$. 

Based on the observation stated above, the contribution of $\alpha$ to the entropy can be expressed as:
\begin{equation*}
    -\eta \cdot \pi_\theta^k(\alpha) \cdot 
    \text{Cov}\!\left(\log \pi_\theta^k(\alpha), C \right) = 0,
\end{equation*}
which indicates that only the retained tokens contribute to the overall entropy.

In other words, we manipulate the number of tokens that can contribute to the entropy by altering the parameter $\varepsilon$.

\subsubsection{Advanced Analysis : Which Type of Tokens Matter Most for Entropy?}\label{analysis}

To understand how individual tokens contribute to the overall entropy, we first revisit the Proposition \ref{covariance equa} established above. In this section, we provide a more precise definition of tokens with low/high probabilities and advantages. It should be noted that in the analysis experiment (Figure~\ref{fig:neg_pos_is_prob}), we adopt the naive REINFORCE algorithm without clipping. Consequently, tokens with high or low advantages are defined according to the sign of their advantage values, i.e., $>0$ for high advantage and $<0$ for low advantage.

\begin{align*}
\mathcal{H}(\pi_\theta^{k+1}|\bm{x},\bm{y_{<t}}) - \mathcal{H}(\pi_\theta^k|\bm{x},\bm{y_{<t}})  
&\approx -\eta \cdot \text{Cov}_{y_t \sim \pi_\theta^k(\cdot|\bm{x},\bm{y_{<t}})} 
\left( \log \pi_\theta^k(y_t), A(y_t) \cdot \mathcal{X}(y_t)+C \right) \\
&= -\eta \cdot \sum_{p=1}^{T}\pi_{\theta}^{k}(y_p|\bm{x},\bm{y_{<t}}) 
\cdot \big(\log \pi_\theta^k(y_p)-\mathbb{E}_{y_i \sim \pi_\theta^k(\cdot|\bm{x},\bm{y_{<t}})}[\log \pi_\theta^k(y_i)]\big) \\
&\quad \cdot \big(A(y_p) \cdot \mathcal{X}(y_p)-\mathbb{E}_{y_i \sim \pi_\theta^k(\cdot|\bm{x},\bm{y_{<t}})}[A(y_i)\cdot \mathcal{X}(y_i)]\big).
\end{align*}

where T is the size of the dictionary.

For convenience, we denote $\mathbb{E}_{y_i}$ as $\mathbb{E}_{y_i\sim \pi_\theta^k(\cdot|\bm{x},\bm{y_{<t}})}$. As only retained tokens contribute to the entropy, we focus only on tokens that are not clipped. We begin by making the following simplification:

\[
\mathbb{E}_{y_i}\!\left(A(y_i)\cdot \mathcal{X}(y_i)\right)
= \mathbb{E}_{y_{\textnormal{clipped}}}\!\left(A(y_i)\cdot 0\right)
+ \mathbb{E}_{y_{\textnormal{retained}}}\!\left(A(y_i)\cdot 1\right)
= \mathbb{E}_{y_{\textnormal{retained}}}\!\left(A(y_i)\right).
\]

So for a selected token $y_s$, its contribution to the overall entropy can be expressed as:

\[
-\eta \cdot \pi_{\theta}(y_s) \cdot (\log \pi_\theta(y_s)-\mathbb{E}_{y_i}(\log \pi_\theta(y_i))) \cdot (A(y_s) -\mathbb{E}_{y_{\textnormal{retained}}}A(y_{\textnormal{retained}})).
\]

Next, we analyze how different types of tokens contribute to the overall entropy. To avoid ambiguity, we first give strict definitions that distinguish between tokens with high/low probabilities and tokens with high/low advantages.

\begin{definition}
    For a token $y_s$, we classify it as follows: 
    \begin{itemize}
        \item \textbf{High advantage:} if 
        \[
        A(y_s) > \mathbb{E}_{y_{\textnormal{retained}}}A(y_{\textnormal{retained}})
        \]
        Otherwise, it is called \emph{low advantage}.
        
        \item \textbf{High probability:} if 
        \[
        \pi_{\theta}(y_s) > \exp\!\left(\mathbb{E}_{y_i}(\log \pi_\theta(y_i)))\right)
        \]
        Otherwise, it is called \emph{low probability}.
    \end{itemize}
\end{definition}

Secondly, we present two propositions that directly follow from the above definitions.

\noindent\textbf{Proposition 3.} 
For a token $y_s$, we have
\[
A(y_s) - \mathbb{E}_{y_{\textnormal{retained}}}A(y_{\textnormal{retained}})
\begin{cases}
> 0, & \text{if $y_s$ is a high-advantage token}, \\[6pt]
< 0, & \text{if $y_s$ is a low-advantage token}.
\end{cases}
\]

\noindent\textbf{Proposition 4.}
For a token $y_s$, we have
\[
\pi_{\theta}(y_s) \cdot (\log \pi_\theta(y_s)-\mathbb{E}_{y_i}(\log \pi_\theta(y_i)))
\begin{cases}
> 0, & \text{if $y_s$ is a high-probability token}, \\[6pt]
< 0, & \text{if $y_s$ is a low-probability token}.
\end{cases}
\]

\begin{proof}
    Let us denote 
    \[
        C = \mathbb{E}_{y_i}(\log \pi_\theta(y_i))),
    \]
    
    which is independent of $y_s$, and let $x = \pi_{\theta}(y_s)$.  As $\pi_{\theta}(y)<1$ for every $y$, $C<0$.\\
    Consider the function 
    \[
        f(x) = x \cdot (\log(x) - C).
    \]  
    Figure~\ref{fig:figure1} illustrates the behavior of this function.  

    \begin{figure}[H]
        \centering
        \includegraphics[width=0.5\textwidth]{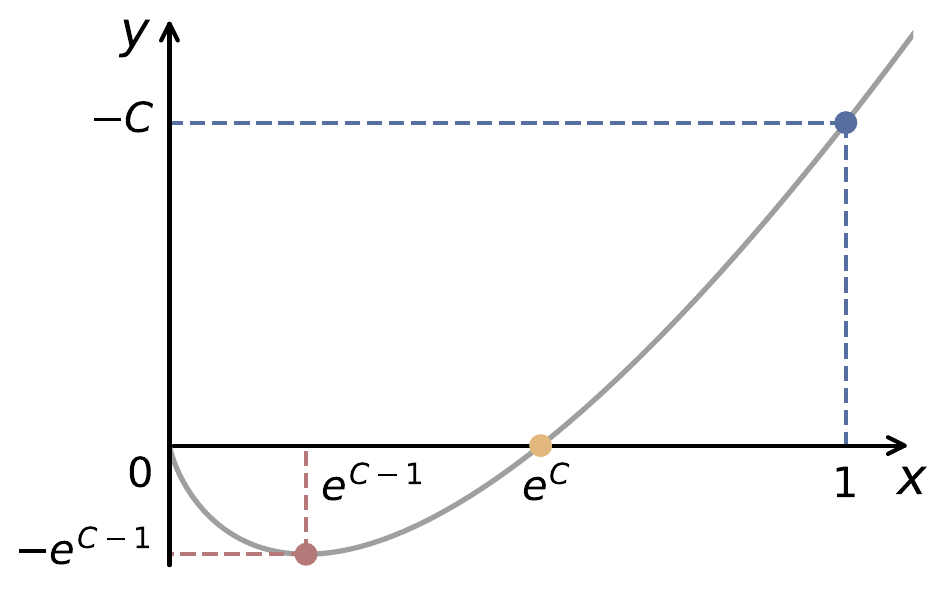}
        \caption{Graph of the function $f(x) = x (\log x - C)$.}
        \label{fig:figure1}
    \end{figure}

    The proposition follows directly from the properties of $f(x)$ as observed in the figure.
\end{proof}

Due to the propositions given above, we have the table below:

\[
\Delta\mathcal{H}(y_s) \approx  -\eta \cdot \underbrace{\pi_{\theta}(y_s) \cdot (\log \pi_\theta(y_s)-\mathbb{E}_{y_i}(\log \pi_\theta(y_i)))}_{\Circled{4}} \cdot \underbrace{(A(y_s) -\mathbb{E}_{y_{\textnormal{retained}}}A(y_{\textnormal{retained}})}_{\Circled{5}}
\]

\begin{table}[H]
    \centering
    \caption{Influence of token characteristics on $\Delta \mathcal{H}(y_s)$. The ``prob'' denotes the probability $\pi_\theta(y_s)$, and the ``adv'' represents the advantage $A(y_s)$.}
    \begin{tabular}{lccc}
\toprule
        \textbf{Token properties} & \textbf{\Circled{4}} & \textbf{\Circled{5}} & $\bm{\Delta\mathcal{H}(y_s)}$ \textbf{($\bm{-\eta \cdot} \textbf{\Circled{4}} \bm{\cdot} \textbf{\Circled{5}}$ )} \\
\midrule
        high prob, high adv & $>0$ & $>0$ & $<0$ \\
        high prob, low adv & $>0$ & $<0$ & $>0$ \\ 
        low prob, high adv & $<0$ & $>0$ & $>0$ \\ 
        low prob, low adv & $<0$ & $<0$ & $<0$ \\ 
\bottomrule
    \end{tabular}
    \label{tab:example_table}
\end{table}

It should be noted that a token $y_s$ \textbf{decreases} the entropy if $\Delta \mathcal{H}(y_s) < 0$, and \textbf{increases} it otherwise.

Therefore, we observe that tokens which are positive with high probabilities and high advantages, or negative with low probabilities and low advantages, contribute to a reduction in the overall entropy. Conversely, positive tokens with high probabilities but low advantages, and negative tokens with high probabilities but low advantages, contribute to an increase in the overall entropy. This observation justifies the statement made in the main part of the thesis.

\end{document}